\documentclass[11pt,letterpaper,onecolumn]{googledeepmind} % logo disabled

\usepackage{amsmath,amsfonts,bm}

\def\eqref#1{equation~\ref{#1}}
\def\1{\bm{1}}

\DeclareMathAlphabet{\mathsfit}{\encodingdefault}{\sfdefault}{m}{sl}
\SetMathAlphabet{\mathsfit}{bold}{\encodingdefault}{\sfdefault}{bx}{n}

\usepackage[authoryear,round]{natbib}
\usepackage{nicefrac}
\definecolor{highlight}{HTML}{F53D7A}
\definecolor{reference}{HTML}{E85D4E}
\definecolor{warmblue}{RGB}{70,110,180}
\hypersetup{
    pdftitle={SimVLA: Zero-Shot Sim-to-Real VLA Learning for Mobile Manipulation},
    pdfauthor={Kyoungin Baik and Youngwoon Lee},
    pdfsubject={Zero-shot sim-to-real vision-language-action learning for mobile manipulation},
    pdfkeywords={vision-language-action, robotics, mobile manipulation, sim-to-real},
    colorlinks=true,
    linkcolor=highlight,
    citecolor=highlight,
    urlcolor=highlight
}
\makeatletter
\newcommand{\srlteaser}{%
    {\centering\projectpagefont \textbf{Project page:} \href{https://kyounginbaik.github.io/simvla/}{\texttt{https://kyounginbaik.github.io/simvla/}}\par}%
    \vspace{4pt}%
    \begin{center}
        \includegraphics[width=0.9\linewidth]{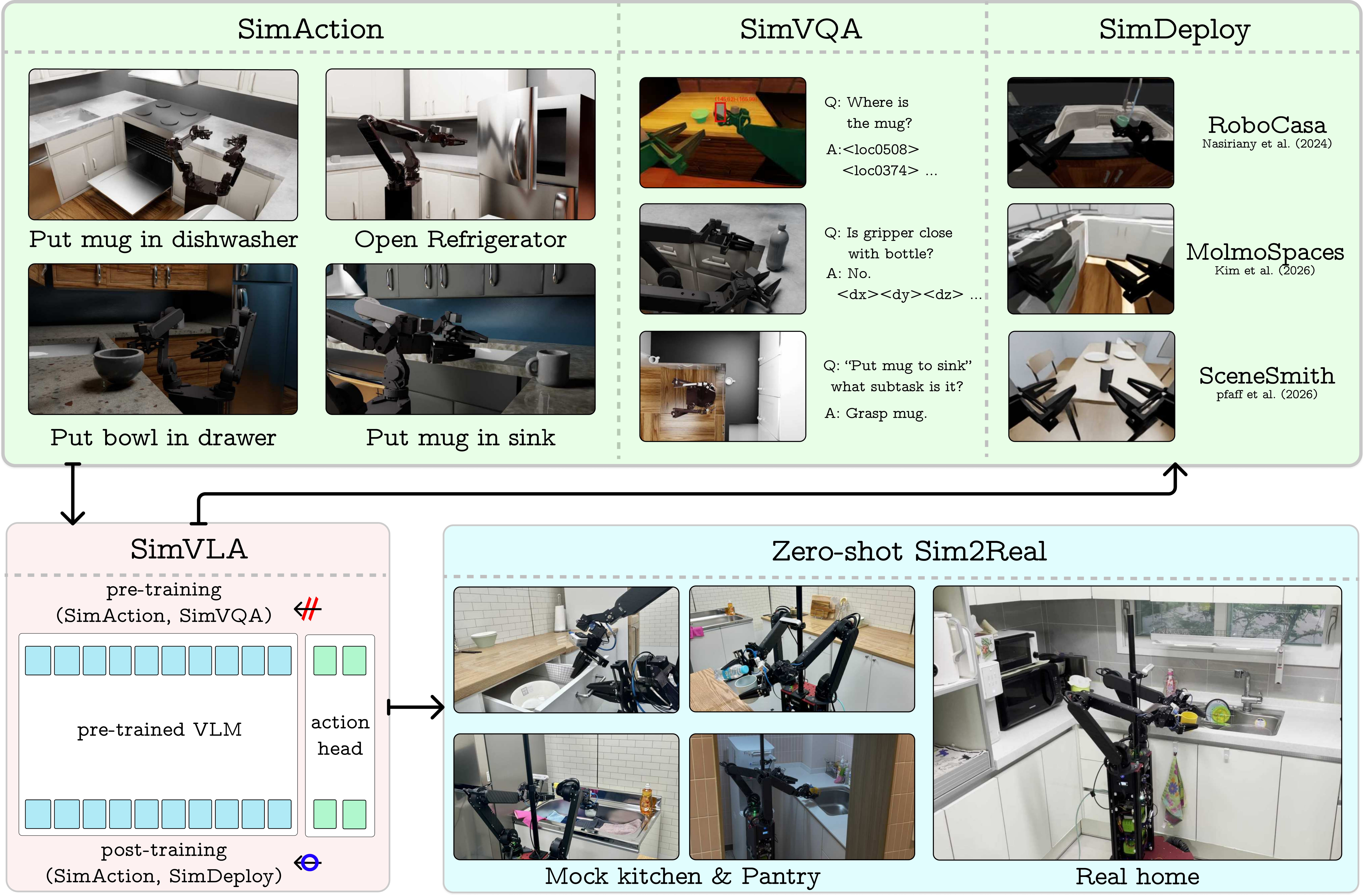}
        \captionof{figure}{SimVLA: Zero-shot sim-to-real framework for mobile manipulation.}
        \label{fig:overview}
    \end{center}
}
\pretocmd{\abscontent}{%
    \par\vspace{-25pt}%
    \noindent{\color{gray}\rule{\linewidth}{1pt}}%
    \par\vspace{10pt}%
    \srlteaser
}{}{}
\makeatother
\usepackage{wrapfig}
\usepackage{float}
\usepackage{multirow}
\usepackage{multicol}
\usepackage{subcaption}
\usepackage{array}
\usepackage{bbm}
\usepackage{pifont}
\usepackage{mathtools}
\usepackage{algorithm}
\usepackage{algpseudocode}
\usepackage{listings}
\lstdefinestyle{srlcode}{
    basicstyle=\ttfamily\footnotesize,
    keywordstyle=\color{srlnavy}\bfseries,
    commentstyle=\color{srlgray}\itshape,
    stringstyle=\color{reference},
    frame=single,
    rulecolor=\color{gray},
    breaklines=true,
    breakindent=2em,
    columns=fullflexible,
    keepspaces=true,
    showstringspaces=false,
    xleftmargin=0pt,
    numbers=none,
    captionpos=b,
}

\tcolorboxenvironment{proposition}{srlbox}
\usepackage[capitalize,noabbrev]{cleveref}
\crefname{assumption}{Assumption}{Assumptions}
\Crefname{assumption}{Assumption}{Assumptions}
\crefname{lstlisting}{Listing}{Listings}
\Crefname{lstlisting}{Listing}{Listings}
\graphicspath{{./}}
\setlist[itemize]{
    topsep=4pt plus 1pt minus 2pt,
    partopsep=1pt plus 0.5pt minus 0.5pt,
    itemsep=2pt plus 1pt minus 0.5pt,
    parsep=2pt plus 1pt minus 0.5pt
}

\renewcommand\Affilfont{\normalfont\fontsize{9}{11.5}\selectfont}
\renewcommand{\absfont}{\normalfont\linespread{1.2}\fontsize{10}{12}\selectfont}
\newcommand{\srlpagenumber}{\normalfont\fontsize{9}{11}\selectfont\thepage}
\fancypagestyle{firststyle}{%
    \fancyhf{}%
    \fancyfoot[C]{\srlpagenumber}%
}
\crefname{equation}{Eq.}{Eqs.}

\newcommand{\xmark}{\ding{55}} % cross mark
\newcommand{\omark}{\ensuremath{\bigcirc}}

\usepackage{xspace}

\title{SimVLA: Zero-Shot Sim-to-Real VLA Learning for Mobile Manipulation}

\author{Kyoungin Baik\textsuperscript{1} and Youngwoon Lee\textsuperscript{2}\\
\Affilfont\textsuperscript{1}Yonsei University , 
\Affilfont\textsuperscript{2}Seoul National University\\
\texttt{dennybaik@yonsei.ac.kr, youngwoon@snu.ac.kr}}
\reportnumber{}

\begin{abstract}
Large-scale, diverse datasets have driven the success of LLMs and VLMs. But VLAs for robotics remain limited by the cost and complexity of real-world data collection. While simulation offers a scalable alternative, its potential for sim-to-real VLA learning in mobile manipulation remains largely underexplored. We introduce \textbf{SimVLA}, an end-to-end framework that trains VLAs \textit{entirely on synthetic simulation data} without teleoperation for mobile manipulation. SimVLA is first pre-trained on two complementary simulation-derived datasets: SimAction, a large-scale robot action dataset spanning 35 diverse mobile manipulation tasks, generated by composing atomic skills, and SimVQA, which leverages privileged simulator state to provide spatial, geometric, and subtask-level visual-language supervision. We further post-train SimVLA on a mixture of SimAction and SimDeploy, a dataset collected from policy rollouts across diverse \textit{simulated} environments. We evaluate SimVLA on tasks including restocking, pouring, and cleaning, and show zero-shot transfer to real-world mobile manipulation, including real home environments. SimVLA outperforms policies trained on 50 in-domain real-world demonstrations, suggesting that simulation can enable scalable sim-to-real mobile manipulation. We further demonstrate the value of multiple complementary forms of supervision for effectively leveraging simulation in VLA training.
\end{abstract}

\begin{document}
\maketitle

\section{Introduction}

Large Language Models (LLMs) and Vision-Language Models (VLMs) have benefited from large-scale, heterogeneous datasets that provide broad coverage over tasks, objects, and environments~\citep{radford2021clip,chen2022pali,alayrac2022flamingo}. Recent Vision-Language-Action (VLA) models extend this data-driven paradigm to robotics by conditioning policies on visual observations and language instructions while predicting robot actions~\citep{Zitkovich2023RT2,octo2023octo,black2024pi0, black2025pi05,kim2025openvla,nvidia2025gr00tn1}. However, unlike web-scale text-image data, robot action data are expensive to collect, embodiment-specific, and limited in task and environmental diversity. This data bottleneck is especially severe for long-horizon mobile manipulation.

Mobile manipulation amplifies the challenges of VLA training in several ways. First, the setting is partially observable~\citep{jauhri2024active,li2025momagen}: a robot may need to act beyond its current field of view. Second, mobility increases the dimensionality of the action space and tightly couples manipulation with navigation~\citep{honerkamp2023n2m2,yang2024harmonic,wu2025moto,shan2026dockanywhere}. For example, a poor base position can cause task failure even when the manipulation policy is otherwise capable. As a result, learning VLAs for mobile manipulation requires broad coverage over scene layouts, viewpoints, and base--arm coordination. Collecting such coverage through real-world teleoperation is difficult~\citep{fu2024mobilealoha}, motivating simulation as a scalable data source if the learned policies can transfer to real environments. Recent work~\citep{deshpande2026molmobot} has taken initial steps toward mobile manipulation, but real-world evaluation remains limited to a single door-opening task and limited performance, highlighting the need for a more effective pipeline.

We introduce \textbf{SimVLA}, a zero-shot sim-to-real framework for training VLA policies on long-horizon mobile manipulation tasks. Unlike prior simulation-based data pipelines that primarily generate or augment action trajectories, SimVLA uses simulation as a source of three complementary forms of supervision: action trajectories, privileged visual-language supervision, and deployment-oriented policy rollouts, as illustrated in \Cref{fig:overview}. First, \textbf{SimAction} provides large-scale, diverse synthetic robot action data spanning 35 mobile manipulation tasks in procedurally generated kitchen environments, without teleoperation. SimAction generates task trajectories by composing atomic skill APIs, allowing the same task structure to be instantiated across diverse layouts, objects, and initial configurations. Second, \textbf{SimVQA} leverages simulation ground-truth to provide spatial, geometric, and subtask-level supervision for VLA co-training. Third, \textbf{SimDeploy} consists of policy rollouts in diverse simulated environments~\citep{robocasa2024,kim2026molmospaces, pfaff2026scenesmith}, exposing the model to deployment-oriented trajectories with broader scene layouts. Together, these data sources allow SimVLA to exploit simulation beyond action generation alone.

SimVLA is trained in two stages. The model is pre-trained on SimAction for action supervision and SimVQA for visual-language supervision. We then roll out the policy in diverse simulated environments to collect SimDeploy, and post-train SimVLA on a mixture of SimAction and SimDeploy, which we find critical for generalization to unseen real-world environments. 

We evaluate SimVLA on long-horizon mobile-manipulation tasks in the real world, including mock kitchens, a pantry, and a real home. Without any real-world data for training, SimVLA transfers to unseen real-world environments. It outperforms a $\pi_{0.5}$ policy fine-tuned on 50 real-world demonstrations in the demonstration environment. It also maintains robust performance in a pantry and a real home, where both baselines degrade sharply. These results suggest that combining action, visual-language, and deployment-oriented supervision in simulation can enable zero-shot sim-to-real mobile manipulation.

\begin{table*}[h]
\centering
\scriptsize
\caption{Comparison of simulation data generation methods. ``V-L sup.'' indicates visual-language supervision beyond action trajectories, such as VQA-style annotations. ``Policy rollouts'' denotes data collected by rolling out a learned policy in simulation.}
\label{tab:sim_datasets}
\resizebox{\textwidth}{!}{%
\begin{tabular}{@{}c c c c c c@{}}
\toprule
Dataset & \# Task & Mobility & V-L sup. & Data generation & Zero-shot Sim-to-Real \\
\midrule
MimicGen \citep{mandlekar2023mimicgen}
& 18 & \omark & \xmark & Teleoperation \& Augmentation & \xmark \\

RoboGen \citep{wang2024robogen}
& 106 & \omark & \xmark & Motion planning \& Optimization \& RL & \xmark \\

SkillMimicGen \citep{Garrett2025skillmimicgen}
& 18 & \xmark & \xmark & Teleoperation \& Augmentation & \xmark \\

DexMimicGen \citep{jiang2025dexmimicgen}
& 9 & \omark & \xmark & Teleoperation \& Augmentation & \omark \\

GraspVLA \citep{deng2025graspvla}
& 1 & \xmark & \omark & Motion planning & \omark \\

RoboTwin 2.0 \citep{chen2026robotwin}
& 50 & \xmark & \xmark & Motion planning & \omark \\

InternData-A1 \citep{tian2025interndataa1}
& 70 & \xmark & \xmark & Motion planning & \omark \\

MoMaGen \citep{li2025momagen}
& 4 & \omark & \xmark & Teleoperation \& Augmentation & \xmark \\

RoboCasa365 \citep{Nasiriany2026robocasa365}
& 365 & \omark & \xmark & Teleoperation \& Augmentation & \xmark \\

MolmoB0T \citep{deshpande2026molmobot}
& 8 & \omark & \xmark & Motion planning & \omark \\

\midrule
\textbf{SimVLA (ours)}
& 35 & \omark & \omark & Motion planning \& Policy rollouts & \omark \\

\bottomrule
\end{tabular}%
}

\end{table*}

\section{Related Work}

\textbf{Data for Training VLAs.} VLA models rely on large-scale, heterogeneous datasets that pair observations, language instructions, and actions producing generalist robot policies~\citep{open_x_embodiment_rt_x_2023, black2024pi0,black2025pi05, nvidia2025gr00tn1}. However, data remains a central bottleneck for VLAs. Robot action data are primarily collected through teleoperation, making it expensive to scale.

\textbf{Sim-to-Real.} Simulation is widely used for robot learning due to its scalable data generation. However, zero-shot sim-to-real remains challenging because of the sim-to-real gap~\citep{li2025momagen, maddukuri2025simandreal, Nasiriany2026robocasa365}. To tackle this, sim-to-real methods often rely on two complementary strategies: system identification~\citep{lin2025simtorealrl,sobanbabu2025SPI} to reduce dynamics mismatch and domain randomization~\citep{singh2024dextrah,chen2026robotwin} to expose policies to a broad range of physical and visual conditions. We combine hybrid system identification, which calibrates both joint-space and task-space behavior, with broad domain randomization.

\textbf{Synthetic Data Generation.} Synthetic robot data is often generated by augmenting a small set of source demonstrations in simulation~\citep{mandlekar2023mimicgen,Garrett2025skillmimicgen, jiang2025dexmimicgen, li2025momagen}. While this improves data quantity, the resulting distributions remain constrained by the source demonstrations. For example, demonstrations for opening a cabinet by pulling may not transfer to cabinets that require different articulated motions, such as rotation. Motion-planning-based approaches address this by generating synthetic data without relying on teleoperation, and recent work has shown strong sim-to-real transfer~\citep{deng2025graspvla,tian2025interndataa1,deshpande2026molmobot}. However, existing approaches largely treat simulation only as a source of action data.

SimVLA treats simulation as a source of multiple complementary forms of supervision. SimAction provides action trajectories generated by composing atomic skill APIs. SimVQA provides spatial, geometric, and subtask-level visual-language supervision derived from privileged simulator state. SimDeploy generates rollout trajectories across diverse simulated environments, exposing the model to policy-induced states. Thus, unlike prior approaches that use simulation mainly to scale action data, SimVLA uses simulation to provide action, visual-language, and deployment-oriented supervision for zero-shot sim-to-real VLA learning.

\textbf{Mobile Manipulation.} Compared to stationary, mobile manipulation remains relatively underexplored despite its importance for real-world tasks. Tasks such as cleaning, restocking, and tidying require not only contact-rich manipulation, but also coordinated movement to reach suitable viewpoints and interaction poses. This makes mobile manipulation fundamentally harder than tabletop manipulation, as policy behavior depends jointly on navigation, base--arm coordination, and partial observability~\citep{honerkamp2023n2m2,lin2025echovla}. Prior work often simplifies mobile manipulation by decoupling navigation and manipulation~\citep{yang2025mobipi}, replaying base trajectories~\citep{Nasiriany2026robocasa365}, or assuming favorable initial poses~\citep{chai2026nm}. While these choices reduce control complexity, they limit the ability to learn coordinated base--arm behaviors~\citep{yang2024harmonic}. Recent work has generated synthetic data for mobile manipulation~\citep{deshpande2026molmobot}, but real-world deployment remains limited in both task diversity and performance. 

In this work, SimVLA targets end-to-end VLA learning under coupled base--arm control across 35 diverse practical tasks and evaluates zero-shot transfer to three unseen real-world environments.

\section{SimVLA: Synthetic Data Generation and VLA Training}
\label{sec:simvla}

SimVLA is a zero-shot sim-to-real framework for training VLA policies \textit{solely from synthetic simulation data}. In this work, we focus on kitchen environments as a representative domain for mobile manipulation tasks. The framework first constructs procedurally generated kitchen scenes (\Cref{sec:proceduralkitchenscenes}). From these scenes, SimVLA derives three complementary data sources: \textbf{SimAction}, a large-scale robot action dataset generated through skill composition and motion planning (\Cref{sec:simaction}); \textbf{SimVQA}, a visual question-answering dataset from privileged simulator state to provide spatial, geometric, and subtask-level supervision (\Cref{sec:simvqa}); and \textbf{SimDeploy}, a deployment-oriented dataset by rolling out the learned policy in diverse simulated environments (\Cref{sec:simdeploy}). The SimVLA model is first pre-trained on SimAction and SimVQA, and then post-trained on a mixture of SimAction and SimDeploy (\Cref{sec:simvla-training}).

\subsection{Procedural Kitchen Scenes}
\label{sec:proceduralkitchenscenes}
Large-scale action data requires diverse environments, objects, and initial configurations. We use Scene Synthesizer~\citep{eppner2025scenesynthesizer} to procedurally generate kitchen layouts, and simulate all scenes in Isaac Lab~\citep{mittal2025isaac}. We instantiate five common kitchen types: island, L-shaped, peninsula, single-wall, and U-shaped layouts. Within each layout type, we randomize furniture heights, furniture and object arrangements, cabinet configurations, and materials. For each object category, we include diverse shape variations from Objaverse~\citep{Deitke2023objaverse} to promote generalization across object geometries. Additional scene and object details are provided in \Cref{sec:kitchen_scene}.

\subsection{SimAction: Synthetic Action Data}
\label{sec:simaction}

SimAction is a large-scale synthetic action dataset for mobile manipulation. Each trajectory contains observations, a language instruction, and continuous robot actions (end-effector poses of two arms and base velocity) for executing long-horizon tasks in procedurally generated kitchen scenes. SimAction comprises approximately 200K trajectories (2,778 hours of robot data) spanning 35 mobile manipulation tasks across 100 procedurally generated houses. The dataset is generated without teleoperation by composing atomic skills into task programs and converting their subtask goals into collision-free robot trajectories through motion planning.

\textbf{Task generation with skill APIs.} Given a generated kitchen scene, a scripted skill API generates subtask-level goal states for robot trajectory generation. The API provides skills for grasping, placing, pouring, boiling, opening, closing, serving, navigating, and searching, which can be composed into reusable long-horizon task specifications. These specifications can then be instantiated across robots, environments, and object configurations to automatically generate demonstrations at scale. For example, put mug in sink decomposes into searching for and navigating to the mug, grasping it, searching for and navigating to the sink, and placing the mug in the sink.

Searching is essential in mobile manipulation, which is inherently partially observable: the robot needs to locate task-relevant objects before interacting with them. The search skill API enables the policy to actively search for task-relevant objects using only visual input at inference time. The detailed logic of the search skill API is described in Section~\ref{sec:skill_api}.

We automatically generate grasp poses using BODex~\citep{chen2025bodex}, while additionally providing a GUI for optional user-defined grasp preferences. The same GUI also allows users to define new tasks by specifying a sequence of atomic skills, enabling scalable task generation without writing low-level control code. A figure of the GUI is provided in \Cref{sec:skill_api}.

\textbf{Trajectory generation with motion planning.} Before generating trajectories, we perform hybrid system identification to reduce dynamics mismatch between simulation and the real robot. Unlike prior work, which often identifies parameters by matching joint-space trajectories~\citep{lin2025simtorealrl}, we additionally constrain end-effector tracking since our policy operates in end-effector space. The full procedure is described in \Cref{alg:r2s_autotune_eef}.

With the identified parameters, we instantiate parallel environments and generate SimAction trajectories by executing motion plans. Given subtask goal states from the skill API, we use cuRobo~\citep{sundaralingam2023curobo} to compute collision-free trajectories. To support embodiment diversity, we generate for three robot embodiments: Anubis~\citep{kang2025anubis}, RB-Y1, and ROBOTIS AI Worker.

\subsection{SimVQA: Simulation-Derived Visual-Language Supervision}
\label{sec:simvqa}

VLAs require visual-language backbones that understand spatial relations, object geometry, subtask progress, and robot-centric coordinates. However, existing sim-to-real VLA pipelines primarily use simulation to generate robot action trajectories~\citep{tian2025interndataa1,deshpande2026molmobot}, leaving much of the privileged information available in simulation unused. SimVQA addresses this gap by converting simulator ground truth into visual question-answering supervision.

For each simulated scene and trajectory, we automatically generate question-answer pairs about object locations, spatial relations, geometric properties, robot state, and subtask progress, yielding approximately 1M SimVQA samples. This provides an intermediate supervision layer between perception and action, encouraging the VLM backbone to encode information relevant for downstream control. Obtaining such spatially grounded annotations from real-world demonstrations would require expensive post-processing~\citep{chenrobo2vlm}. In simulation, the same annotations are available from ground-truth state at no additional cost. All VQA templates are provided in \Cref{App:SimVQA}, \Cref{tab:vqa_prompt_templates}.

\subsection{SimDeploy: Deployment-Oriented Rollout Data}
\label{sec:simdeploy}

Although SimAction provides scalable robot action data, scripted motion-planning trajectories do not fully capture the states encountered by a learned policy at deployment time. Small prediction errors can lead to changed viewpoints, imperfect base poses, and missed grasps. To expose the model to such policy-induced states, we construct SimDeploy by rolling out the SimVLA policy zero-shot across diverse simulated environments, e.g., RoboCasa~\citep{robocasa2024}, MolmoSpaces~\citep{kim2026molmospaces}, and SceneSmith~\citep{pfaff2026scenesmith}, collecting 22K trajectories (460 hours). Unlike SimAction, which is generated from scripted skill compositions and motion planning, SimDeploy captures the distribution shift induced by the learned policy itself, including diverse viewpoints and scene layouts. We find that post-training on SimDeploy improves robustness and generalization to unseen real-world environments.

\begin{figure*}[t]
  \centering
  \includegraphics[width=0.9\linewidth]{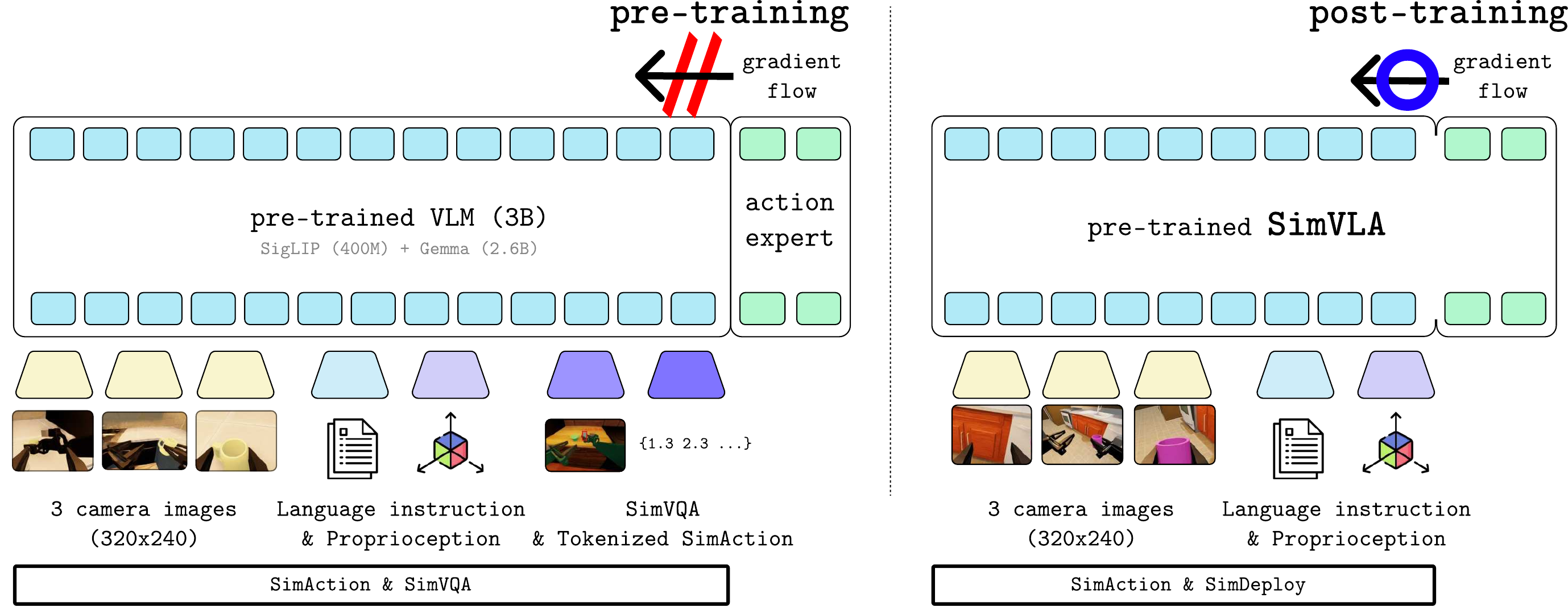}
  \caption{Overview of SimVLA Training. During pre-training, SimVLA jointly optimizes continuous action prediction, tokenized action prediction, and SimVQA supervision. During post-training, SimVLA is trained on a mixture of SimAction and deployment trajectories from SimDeploy.}
  \label{fig:SimVLA}

\end{figure*}

\subsection{SimVLA: Two-Stage VLA Training}
\label{sec:simvla-training}

SimVLA follows a two-stage training pipeline built on a pre-trained VLM~\citep{beyer2024paligemma} and a flow-matching action expert, following the model structure of $\pi_{0.5}$~\citep{black2025pi05}. During pre-training, SimVLA jointly optimizes three simulation-derived objectives in a single forward pass: flow matching on continuous actions from SimAction, autoregressive prediction of tokenized actions~\citep{pertsch2025fast} from SimAction, and autoregressive prediction of SimVQA answers. This stage aligns the VLM backbone with robot action while preserving its visual-language representation. To prevent unstable gradients from the randomly initialized action head from degrading the pre-trained VLM backbone, we stop gradients from the action head into the backbone during this stage~\citep{driess2025knowledge}. 

During post-training, SimVLA optimizes the flow-matching objective on a mixture of SimAction and SimDeploy. Since the action head has been warmed up during pre-training, we allow gradients to update the VLM backbone in this stage. We deliberately select $\pi_{0.5}$-style architecture so that the experiments isolate the effect of the proposed simulation data and supervision. Detailed training objectives and configurations are provided in \Cref{sec:training_details}.

\section{Experiments}
\label{sec:experiments}
\subsection{Simulation Experiments}
We compare SimVLA with MolmoB0T~\citep{deshpande2026molmobot} on RB-Y1 in procedurally generated unseen SimVLA kitchens and MolmoSpaces~\citep{kim2026molmospaces}, both evaluated zero-shot. We additionally evaluate eight long-horizon tasks using the Anubis robot~\citep{kang2025anubis} (\Cref{tab:long_horizon_sim}).  

\subsection{Real-World Experiments}
\begin{figure*}[t]
        \vspace{2pt}
      \centering
      \captionsetup[subfigure]{skip=2pt,font=small}

      \begin{subfigure}[b]{0.3\textwidth}
          \includegraphics[width=\linewidth]{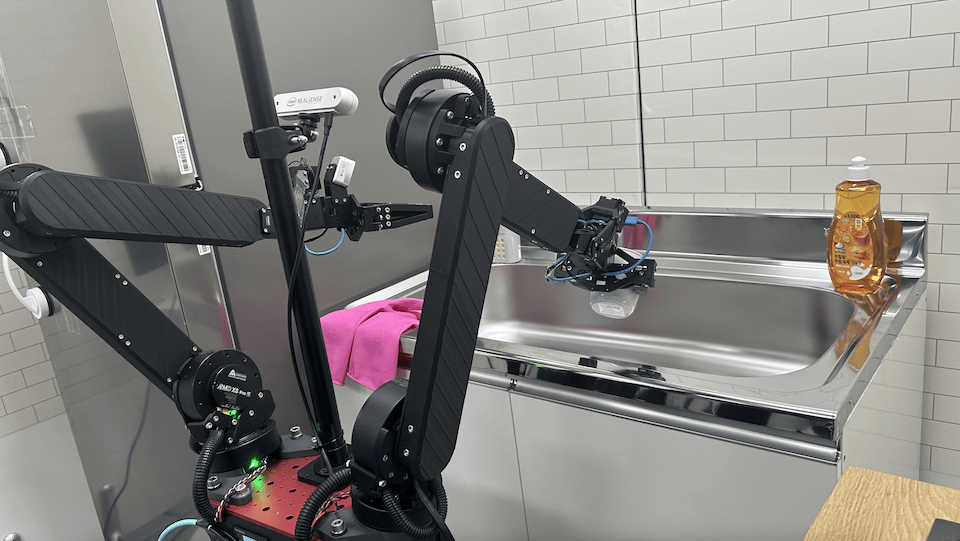}
          \caption{Put mug in sink}
          \label{fig:task_mug_sink}
      \end{subfigure}
      \hfill
      \begin{subfigure}[b]{0.3\textwidth}
          \includegraphics[width=\linewidth]{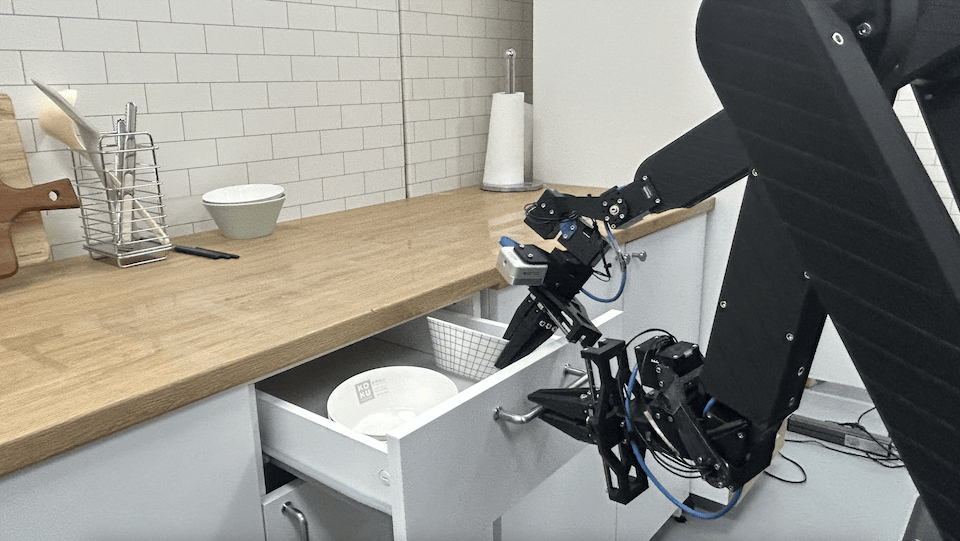}
          \caption{Put bowl in drawer}
          \label{fig:task_bowl_drawer}
      \end{subfigure}
      \hfill
        \begin{subfigure}[b]{0.3\textwidth}
          \includegraphics[width=\linewidth]{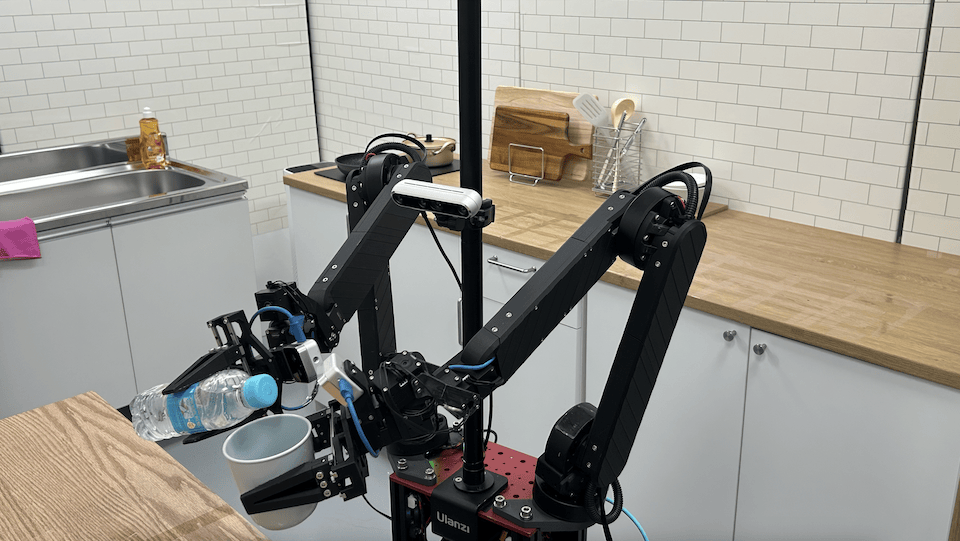}
          \caption{Pour bottle to mug}
          \label{fig:task_pour}
      \end{subfigure}

      \vspace{4pt}

      \begin{subfigure}[b]{0.3\textwidth}
          \includegraphics[width=\linewidth]{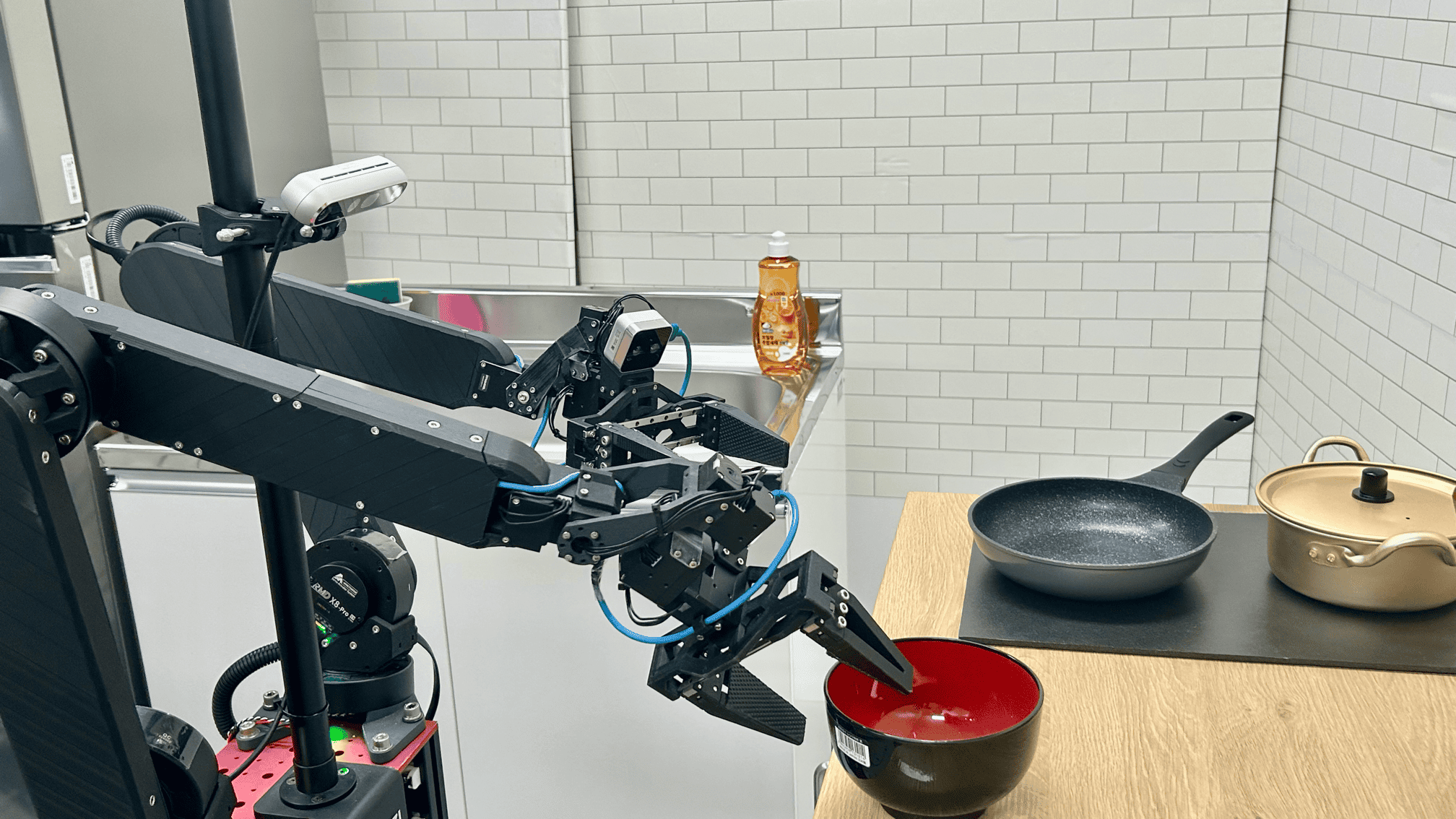}
          \caption{Put bowl in sink}
          \label{fig:task_bowl_sink}
      \end{subfigure}
        \hspace{0.05\textwidth}
                \begin{subfigure}[b]{0.3\textwidth}
          \includegraphics[width=\linewidth]{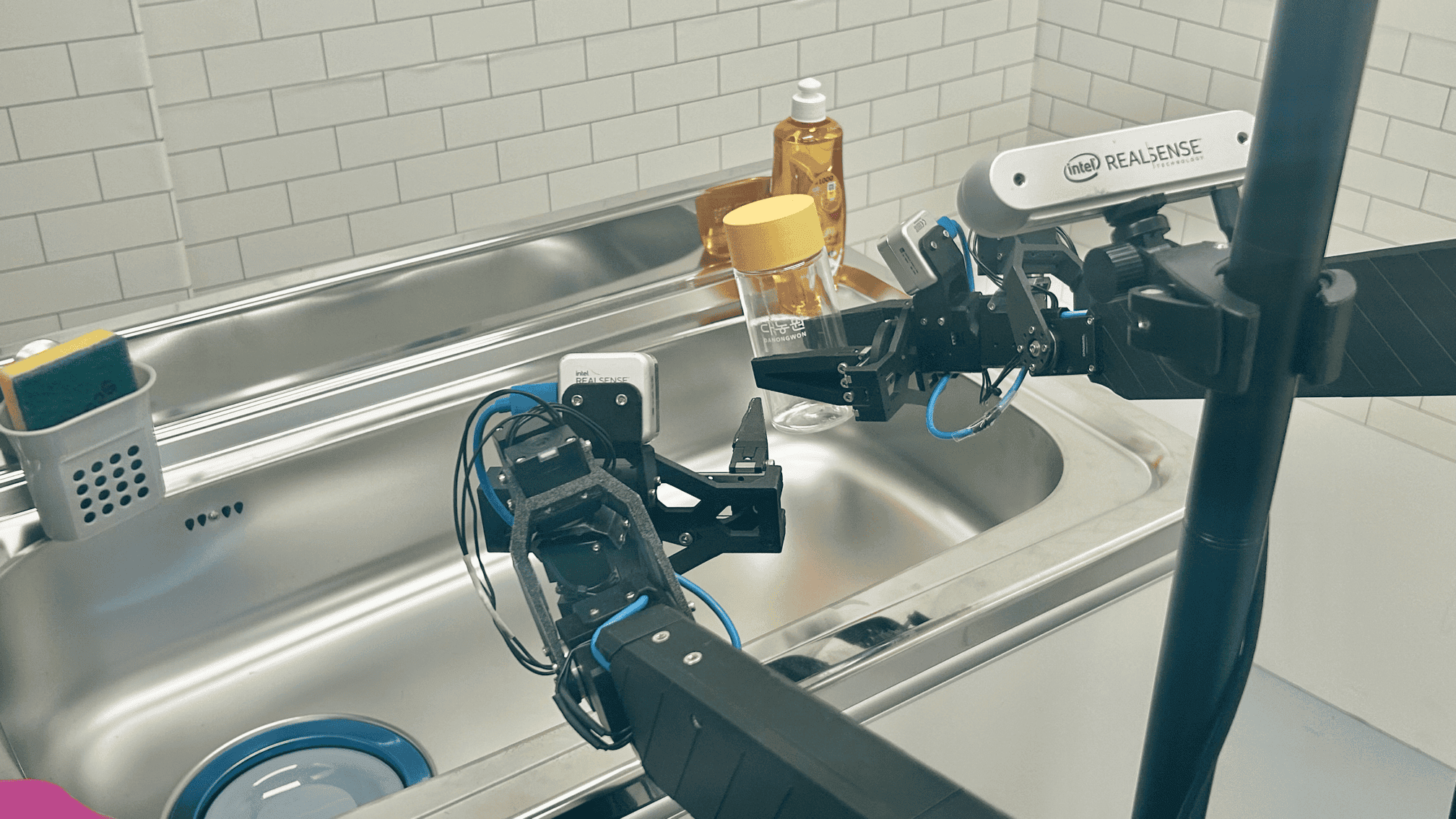}
          \caption{Put bottle in sink}
          \label{fig:task_bottle_sink}
        \end{subfigure}

      \caption{Five long-horizon mobile manipulation tasks used for real-world evaluation.}
      \label{fig:all_tasks}
      
  \end{figure*}

We evaluate SimVLA on five long-horizon mobile manipulation tasks: \textit{put mug / bowl / bottle in sink}, \textit{put bowl in drawer}, and \textit{pour bottle to mug} (\Cref{fig:all_tasks}). These tasks cover distinct challenges, including navigation to receptacles, articulated-object interaction, base--arm coordination, and bimanual pouring. All real-world experiments use Anubis, a bimanual mobile manipulator with two 6-DoF arms, parallel grippers, and a 3-wheel omnidirectional base.

\begin{figure}[h]
\vspace{8pt} 
    \centering
    \captionsetup[subfigure]{skip=2pt}

    % ---- Row 1: evaluation environments ----
    \begin{subfigure}[b]{0.32\textwidth}
        \includegraphics[width=\linewidth]{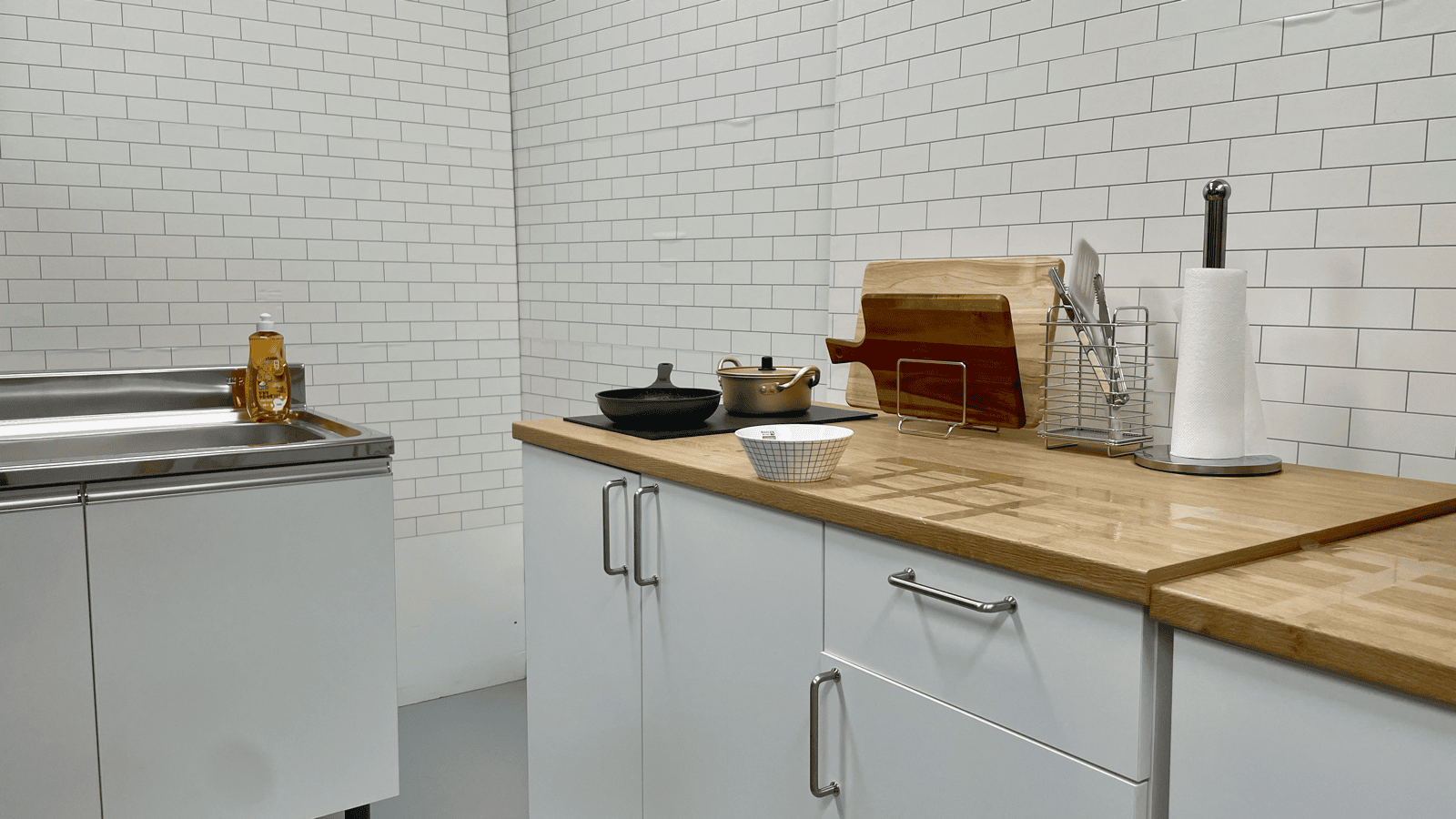}
        \caption{Mock kitchen}
        \label{fig:env_mock}
    \end{subfigure}
    \hfill
    \begin{subfigure}[b]{0.32\textwidth}
        \includegraphics[width=\linewidth]{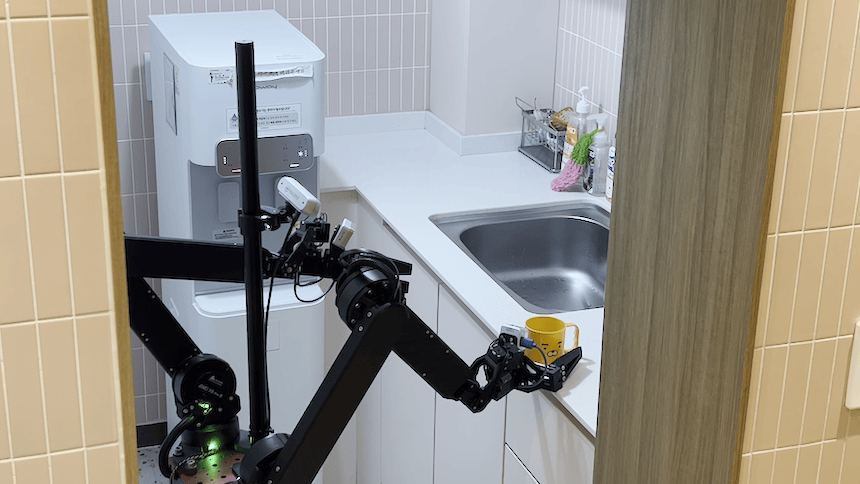}
        \caption{Pantry}
        \label{fig:env_pantry}
    \end{subfigure}
    \hfill
    \begin{subfigure}[b]{0.32\textwidth}
        \includegraphics[width=\linewidth]{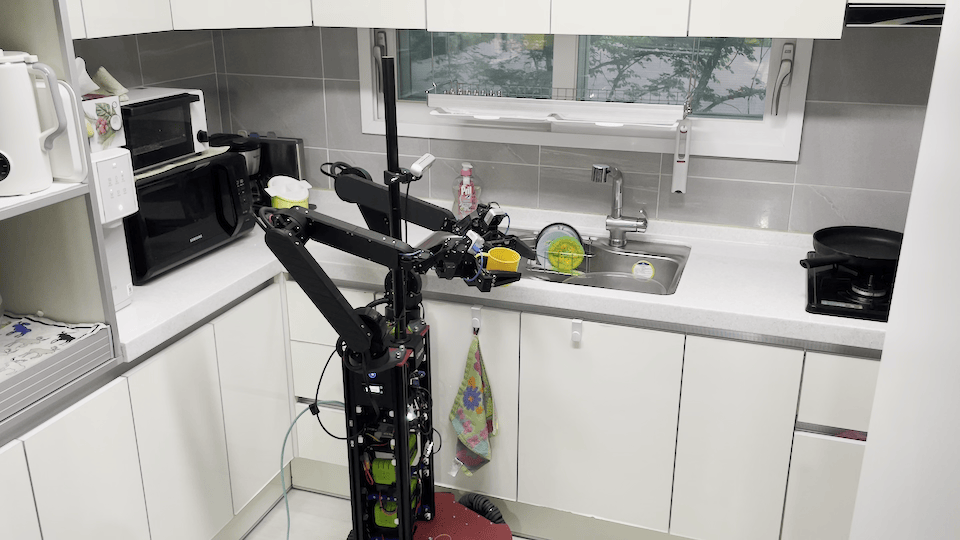}
        \caption{Real home}
        \label{fig:env_home}
    \end{subfigure}

    \vspace{6pt}

    % ---- Row 2: mock kitchen evaluation conditions ----
    \begin{subfigure}[t]{0.32\textwidth}
        \includegraphics[width=\linewidth]{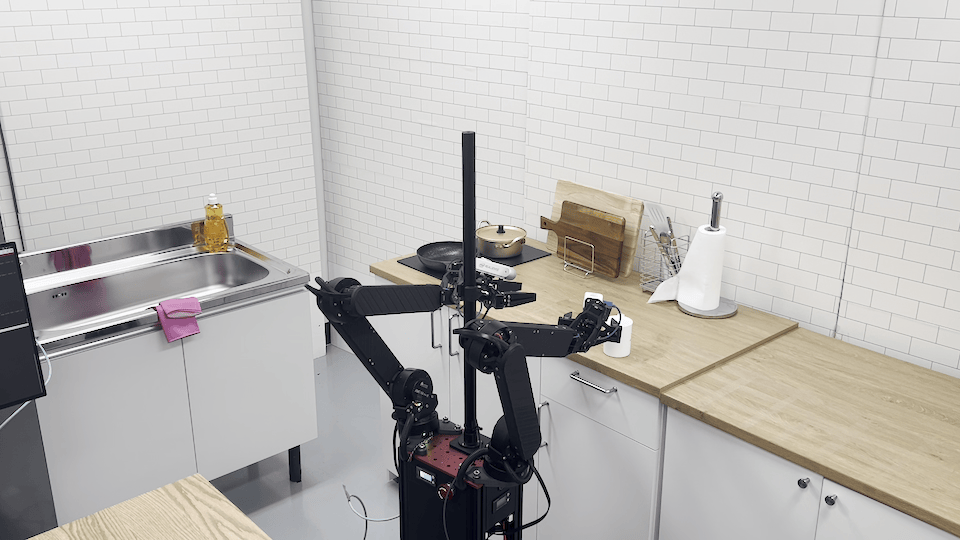}
        \caption{In-Distribution}
        \label{app:3d}
    \end{subfigure}
    \hfill
    \begin{subfigure}[t]{0.32\textwidth}
        \includegraphics[width=\linewidth]{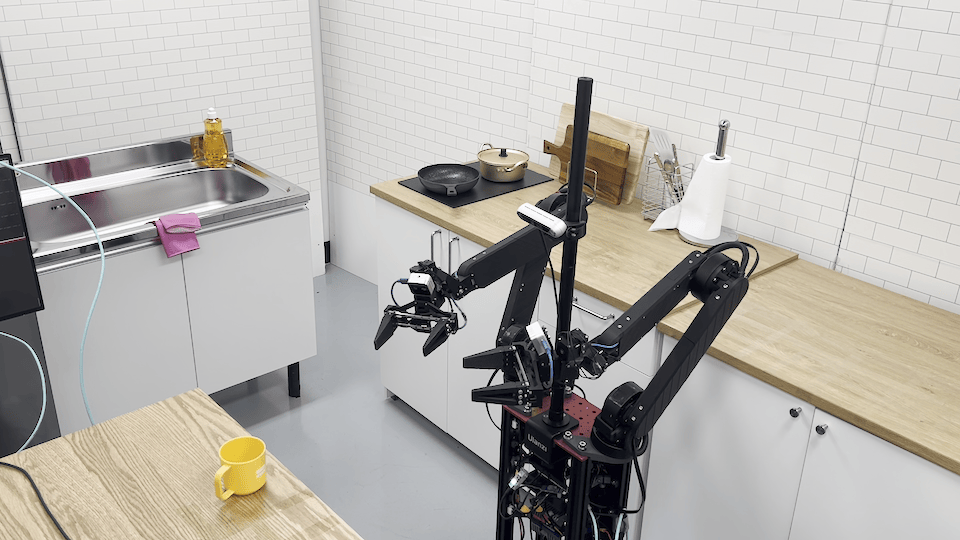}
        \caption{OOD-pos}
        \label{app:3e}
    \end{subfigure}
    \hfill
    \begin{subfigure}[t]{0.32\textwidth}
        \includegraphics[width=\linewidth]{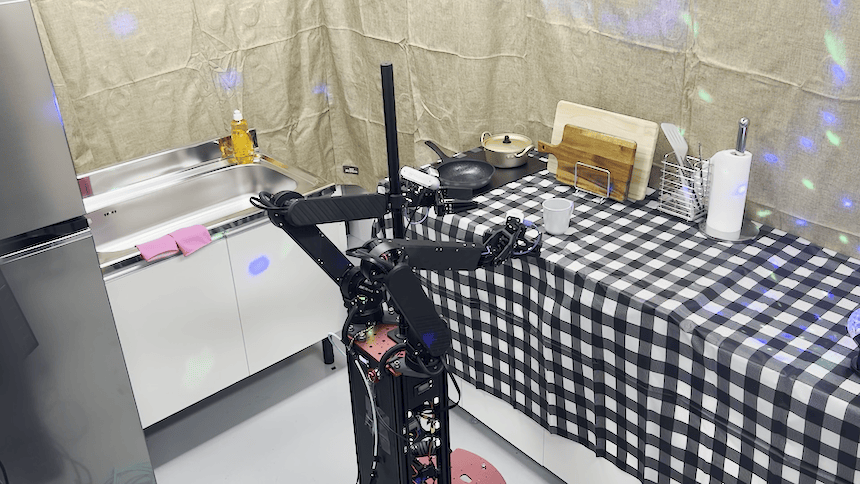}
        \caption{OOD-vis}
        \label{app:3f}
    \end{subfigure}

    \caption{Real-world evaluation setup. \textbf{Top:} three evaluation environments: (\subref{fig:env_mock}) mock kitchen, (\subref{fig:env_pantry}) pantry, and (\subref{fig:env_home}) real home. \textbf{Bottom:} mock kitchen evaluation conditions: (\subref{app:3d}) In-Distribution, (\subref{app:3e}) OOD-pos, and (\subref{app:3f}) OOD-vis.}
    \label{fig:all_envs}
    \label{fig:mock_conditions}
\end{figure}

We evaluate in three real-world environments: a mock kitchen, a pantry, and a real home (\Cref{fig:all_envs}). In the mock kitchen, we consider three conditions: In-Distribution (ID), where data were collected for baseline models; OOD-pos, where objects and furniture are repositioned; and OOD-vis, where backgrounds and lighting are randomized (\Cref{fig:mock_conditions}). 

For baseline methods, we fine-tune $\pi_{0.5}$ on 50 or 200 mock kitchen data per task. All methods share the same observation/action spaces, prompts, and trial protocol. We report task progress, i.e., the percentage of predefined subgoals completed. Each task–policy pair is evaluated over 20 trials in the mock kitchen and pantry and 10 trials in the real home, for a total of 2,120 rollouts. Detailed definitions of subgoals and task progress, along with binary success rates, are provided in \Cref{Task progress description}.

\begin{figure*}[t]
  \centering
  \includegraphics[width=0.9\linewidth]{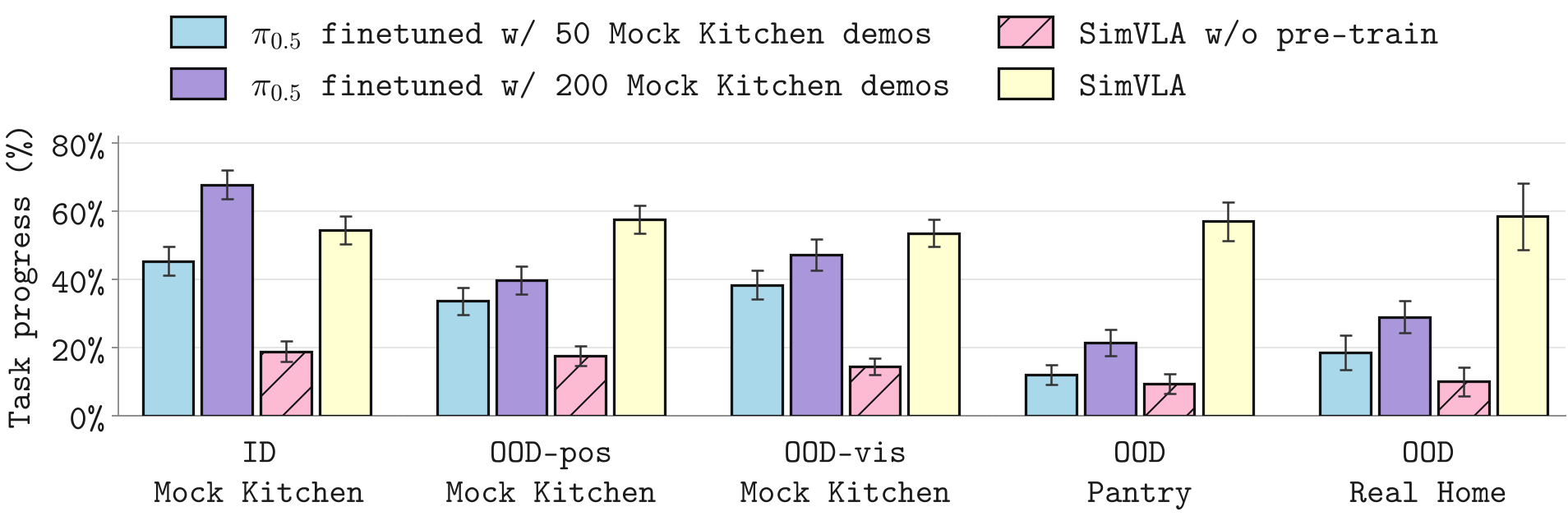}
    \caption{Averaged task progress in real-world environments. Mock kitchen includes all five tasks; pantry includes \textit{put mug / bowl / bottle in sink}; real home includes \textit{put mug in sink} / \textit{pour bottle to mug}. SimVLA maintains robust performance in OOD environments, while the fine-tuned policy degrades substantially. Removing pre-training leads to poor performance, demonstrating its importance for robust generalization.}
    \label{fig:policy_evaluation}
\end{figure*}

\subsection{Can SimVLA Match Policies Trained with In-Domain Demonstrations?}

\textbf{Yes, in our evaluated settings.} We compare SimVLA to $\pi_{0.5}$ fine-tuned on 50 or 200 in-domain real-world demonstrations per task. SimVLA is trained entirely in simulation. As shown in \Cref{fig:policy_evaluation}, SimVLA achieves 54.4\% average task progress, outperforming the $\pi_{0.5}$ policy fine-tuned on 50 in-domain real-world demonstrations in our evaluated settings. Under OOD-pos and OOD-vis, in-domain baselines degrade substantially, and increasing real-world demonstrations from 50 to 200 gives limited gain. This suggests that additional in-domain data alone may not provide sufficient coverage for robust mobile manipulation in OOD settings, while SimVLA benefits from broader synthetic variation.

\subsection{Does SimVLA Need Pre-Training?}
\label{sec:Q1}
\textbf{Yes.}
We compare SimVLA with a variant that skips pre-training and is trained only with the post-training objective on the same SimAction--SimDeploy mixture. As shown in \Cref{fig:policy_evaluation}, removing pre-training substantially degrades performance. This finding is consistent with prior work that leverages large-scale real-world robot data and multimodal web data for pre-training. In contrast, SimVLA obtains its pre-training data, SimAction/SimVQA, entirely from simulation.

Although $\pi_{0.5}$~\citep{black2025pi05} has demonstrated strong generalization to unseen environments, adapting the pre-trained checkpoint to a new task or robot embodiment typically requires additional data collection and fine-tuning. Because these data are collected in a particular environment, the fine-tuned policy can exhibit substantial performance degradation under environment changes. In our experiments, the $\pi_{0.5}$ model fine-tuned with 200 real-world demonstrations achieves 67.8\% task progress in mock kitchen-ID, but only 21.3\% and 29\% in pantry and real home, respectively. In contrast, SimVLA maintains consistent performance across all three real-world environments.

\newpage

\begin{figure*}[t]
  \centering
  \includegraphics[width=0.9\linewidth]{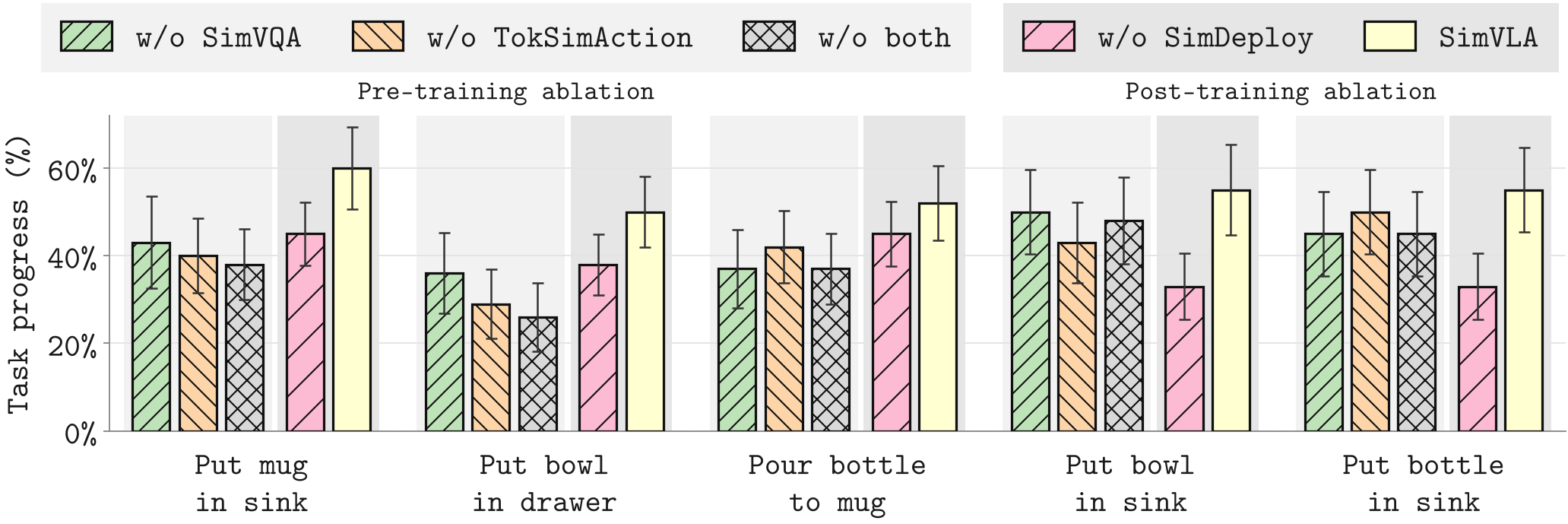}
    \caption{Averaged task progress across tasks in mock kitchen-ID. Removing either SimVQA or Tokenized SimAction degrades performance across all tasks. SimDeploy yields task-dependent gains in zero-shot deployment performance.}
    \label{fig:simvqa_ablation}
\end{figure*}

\subsection{Pre-training Data Ablation}
\label{sec:simvqa_ablation}
\vspace{-0.1em}
We compare SimVLA with variants that remove SimVQA, TokSimAction, or both during pre-training to evaluate their contributions. As shown in \Cref{fig:simvqa_ablation}, removing either component consistently degrades performance across all five tasks, with the full model achieving the best results. Qualitatively, the model pre-trained without SimVQA often fails due to imprecise grasping, suggesting that SimVQA provides valuable spatial supervision for downstream manipulation.

To further investigate this effect, we evaluate SimVLA on 1{,}369 held-out SimVQA pairs. As shown in \Cref{tab:ablation_visual_reasoning}, the full model achieves a median 3D position error of only $0.06$\,m and a cross-view IoU of $0.61$, substantially outperforming the ablated variants. These results indicate that SimVQA pre-training provides the model with accurate spatial reasoning capabilities beyond action prediction.

\subsection{Post-training Data Ablation}

We compare SimVLA against a variant without SimDeploy during post-training in \Cref{fig:simvqa_ablation}. SimDeploy substantially improves \textit{put mug / bowl / bottle in sink} and \textit{put bowl in drawer} as it adds deployment trajectories with diverse layouts, viewpoints, and retry behaviors. However, it gives relatively smaller gains on \textit{pour bottle to mug}. This is consistent with the variation required by each task: pouring mainly requires object-level generalization, already covered by SimAction randomization, whereas sink and drawer tasks require broader scene-level variation.

\newpage

\begin{figure}[t]
    \centering
    \begin{subfigure}{0.45\textwidth}
        \includegraphics[width=\linewidth]{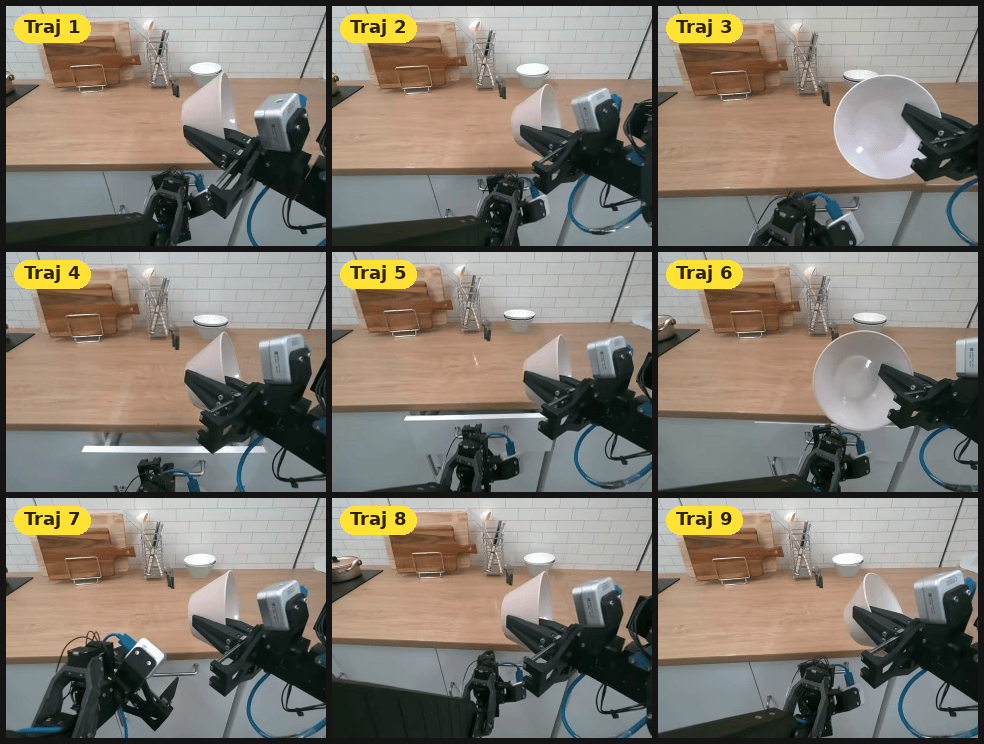}
        \caption{Real-world teleoperation demonstrations}
        \label{real world teleop demo}
    \end{subfigure}
    \hfill
    \begin{subfigure}{0.45\textwidth}
        \includegraphics[width=\linewidth]{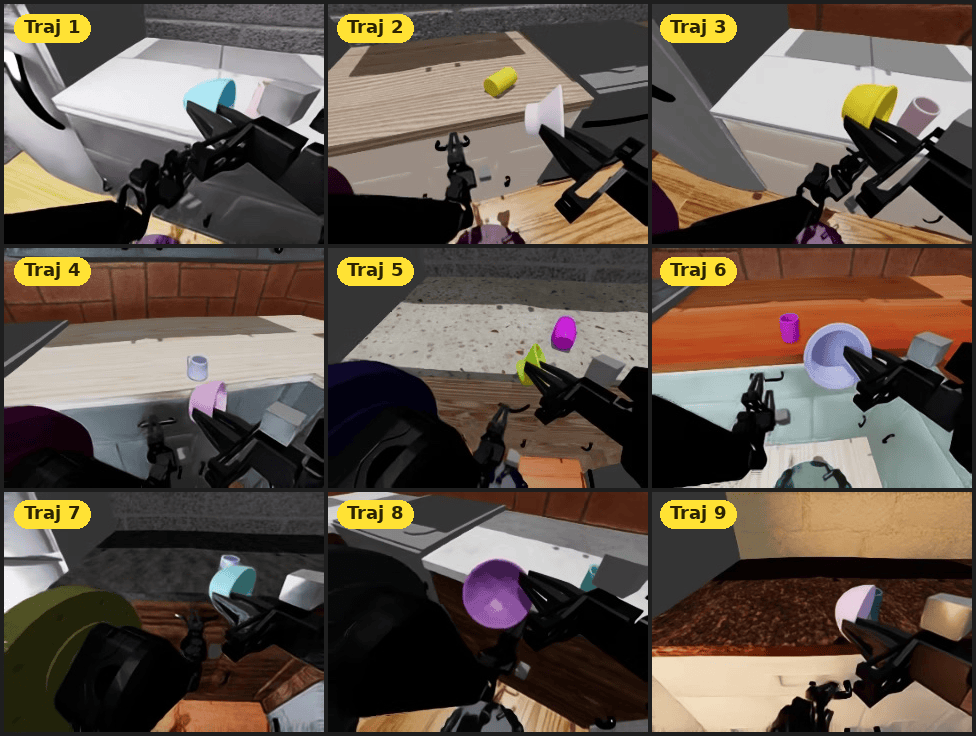}
        \caption{Simulation demonstrations}
        \label{simulation demo}
    \end{subfigure}
    \caption{Sampled drawer-grasping frames from real-world teleoperation and simulation. Teleoperation trajectories are visually homogeneous, while simulation exhibits greater diversity.}
\end{figure}

\subsection{What Makes Mobile Manipulation Hard in OOD Situations?}

\textbf{Limited trajectory coverage.} 
Teleoperated demonstrations often cover narrow viewpoints and approach trajectories, especially for precise interactions such as drawer-handle grasping (\Cref{real world teleop demo}). With action chunking, small errors can push the policy into OOD states; for example, slight over-rotation in the \textit{put bowl in drawer} can misalign the gripper with the handle and cause grasp failure. As a result, in mock kitchen OOD-pos (\Cref{app:3e}), policies trained on real-world demonstrations mostly failed to grasp the handle. In contrast, SimVLA generates more diverse viewpoints and spatial configurations through subgoal domain randomization (\Cref{simulation demo}), resulting in more robust performance. This suggests that limited trajectory coverage is a key bottleneck for OOD generalization, and that large-scale simulation can mitigate this limitation.

\subsection{How does SimVLA compare to existing zero-shot sim-to-real mobile manipulation baselines?}

\begin{wraptable}[17]{r}{0.46\textwidth}
  \vspace{-1.1\baselineskip}
  \centering\small
  \caption{\textbf{Zero-shot mobile manipulation.} Success rate (\%); \textit{Pick \{mug, bottle, apple\}}, \textit{PnP \{apple in bowl, mug next to bowl\}}. SimVLA is post-trained without MolmoSpaces SimDeploy. SimVLA remains robust under partial observability. None of the environments are included in the training dataset.}
  \label{tab:molmobot_vs_simvla}
  \begin{tabular}{@{}lcc@{}}
  \toprule
  & MolmoB0T & SimVLA \\
  \midrule
  \multicolumn{3}{@{}l}{\emph{Pick}} \\
  \quad MolmoSpaces & 36.3\,$\pm$\,2.8 & \textbf{71.0\,$\pm$\,2.9} \\
  \quad SimVLA kitchens & 9.7\,$\pm$\,1.7 & \textbf{91.2\,$\pm$\,2.8} \\
  \midrule
  \multicolumn{3}{@{}l}{\emph{PnP}} \\
  \quad MolmoSpaces & 1.0\,$\pm$\,0.7 & \textbf{49.0\,$\pm$\,3.4} \\
  \quad SimVLA kitchens & 0.5\,$\pm$\,0.5 & \textbf{79.5\,$\pm$\,3.1} \\
  \midrule
  \multicolumn{3}{@{}l}{\emph{Partial observation}} \\
  \quad MolmoSpaces & 0.0\,$\pm$\,0.0 & \textbf{51.8\,$\pm$\,3.1} \\
  \quad SimVLA kitchens & 0.0\,$\pm$\,0.0 & \textbf{81.0\,$\pm$\,2.9} \\
  \bottomrule
  \end{tabular}
  \vspace{-1.2\baselineskip}
\end{wraptable}

\textbf{SimVLA enables more generalizable mobile manipulation.} We compare SimVLA with MolmoB0T using the RB-Y1 robot across five held-out SimVLA kitchens and five MolmoSpaces environments. We evaluate three picking and two pick-and-place tasks with 200 trials per task. While MolmoB0T's performance degrades substantially in SimVLA kitchens, SimVLA maintains stable performance on MolmoSpaces, demonstrating stronger zero-shot transfer.

We also evaluate mobile manipulation when targets are not visible from the initial viewpoint. MolmoB0T achieves 0\% success across all tasks, whereas SimVLA succeeds through learned active vision behaviors. Together, these results demonstrate that SimVLA achieves stronger zero-shot mobile manipulation, generalizing to unseen environments and locating targets under partial observability.

\section{Conclusion}

We presented SimVLA, a zero-shot sim-to-real framework that trains VLAs entirely from simulation data for mobile manipulation. By combining procedural kitchen scene generation, hybrid system identification, domain randomization, and SimVQA supervision, SimVLA produces large-scale, diverse synthetic training data without teleoperation. SimDeploy data collected through policy rollouts across simulation environments further reduces the distribution gap by exposing the policy to scene-level diversity and policy-induced states beyond procedural demonstrations. Experiments across five tasks and three real-world environments show that SimVLA, trained purely in simulation, outperforms policies trained on 50 in-domain real-world data in our evaluation settings. SimVLA also transfers to practical settings, including a pantry and a real home, without environment-specific data. 
Comparison with a current sim-to-real VLA baseline demonstrates SimVLA's generalization across mobile manipulation settings. Furthermore, SimVLA demonstrates that simulation can serve not only as a source of action trajectories, but also as a source of visual-language supervision and deployment-oriented rollout data for zero-shot sim-to-real VLA learning in mobile manipulation.

\section{Limitations \& Future Work}
\label{sec:limitations}

Although SimVLA demonstrates zero-shot sim-to-real transfer for mobile manipulation, it has several limitations. First, we do not isolate the contributions of individual domain randomization factors. Second, our tasks focus on rigid-object manipulation in kitchen-like environments; extending to deformable objects, dexterous manipulation, and broader household settings remains future work. Third, since SimVLA is a simulation-based data generation framework rather than a cross-embodiment transfer method or a new VLA architecture, our evaluation is limited to a subset of embodiments and architectures. Broader evaluation would be needed to assess its generality. Finally, SimAction generation relies on human-defined skill APIs; future work could leverage LLMs to expand skills as code-based policies, enabling a more autonomous pipeline.

\section{Acknowledgements}
\label{sec:Acknowledgements}
This work was supported in part by the National Research Foundation of Korea (NRF) grants funded by the Korean government (MSIT) (RS-2024-00333634, RS-2025-25396144). LG Electronics provided the Anubis robot, which was used for the experiments. The authors would like to thank Changwoo Yoo for the assistance with the preliminary experiments and real-world data collection.

\bibliography{references}
\bibliographystyle{plainnat}

\newpage

\appendix

\section{Kitchen Scene}
\label{sec:kitchen_scene}

\begin{figure}[h]
  \centering
  \includegraphics[width=1\linewidth]{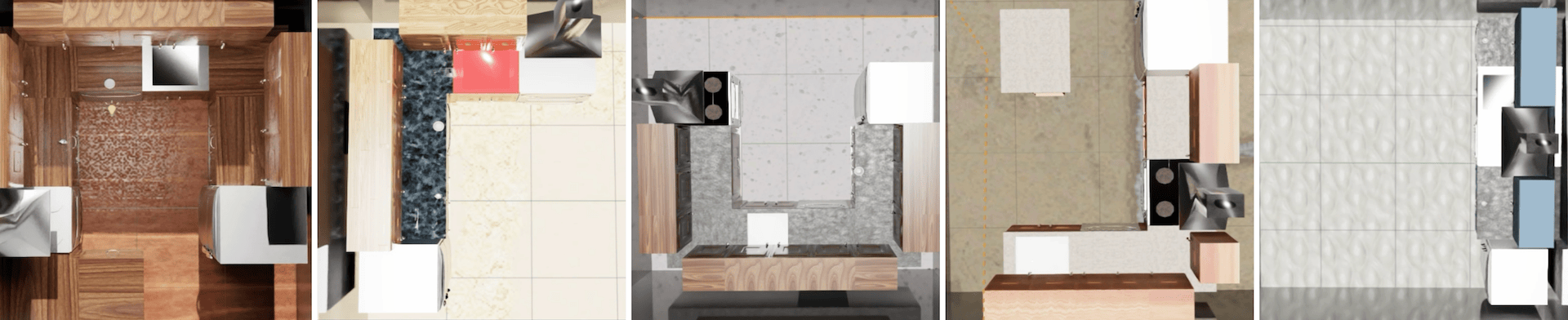}
  \caption{The five kitchen layout types used in SimVLA: peninsula, L-shaped, U-shaped, island, and single-wall. Each layout is procedurally generated with randomized furniture configurations, object placements, and surface materials.}
  \label{fig:kitchen_layout}
  \vspace{-1em}
\end{figure}

\paragraph{Kitchen layout.} To promote structural and visual diversity in our simulation environments, we use five common kitchen layouts from Scene Synthesizer~\citep{eppner2025scenesynthesizer}: island, L-shaped, peninsula, single-wall, and U-shaped, as illustrated in Figure~\ref{fig:kitchen_layout}. Each layout type represents a distinct spatial configuration commonly found in real-world kitchens. Each layout presents distinct navigation and manipulation challenges: \textit{island} layouts require navigation around a central counter, \textit{L-shaped} layouts create challenging corners, \textit{peninsula} layouts form partially enclosed workspaces, \textit{single-wall} layouts provide a compact linear workspace, and \textit{U-shaped} layouts require precise base positioning across multiple surfaces. Within each layout, furniture heights, object arrangements, cabinet configurations, and surface materials are randomized to increase visual and geometric diversity.

\begin{wrapfigure}[11]{r}{0.4\textwidth}
  \vspace{-1em}
  \centering
  \includegraphics[width=0.9\linewidth]{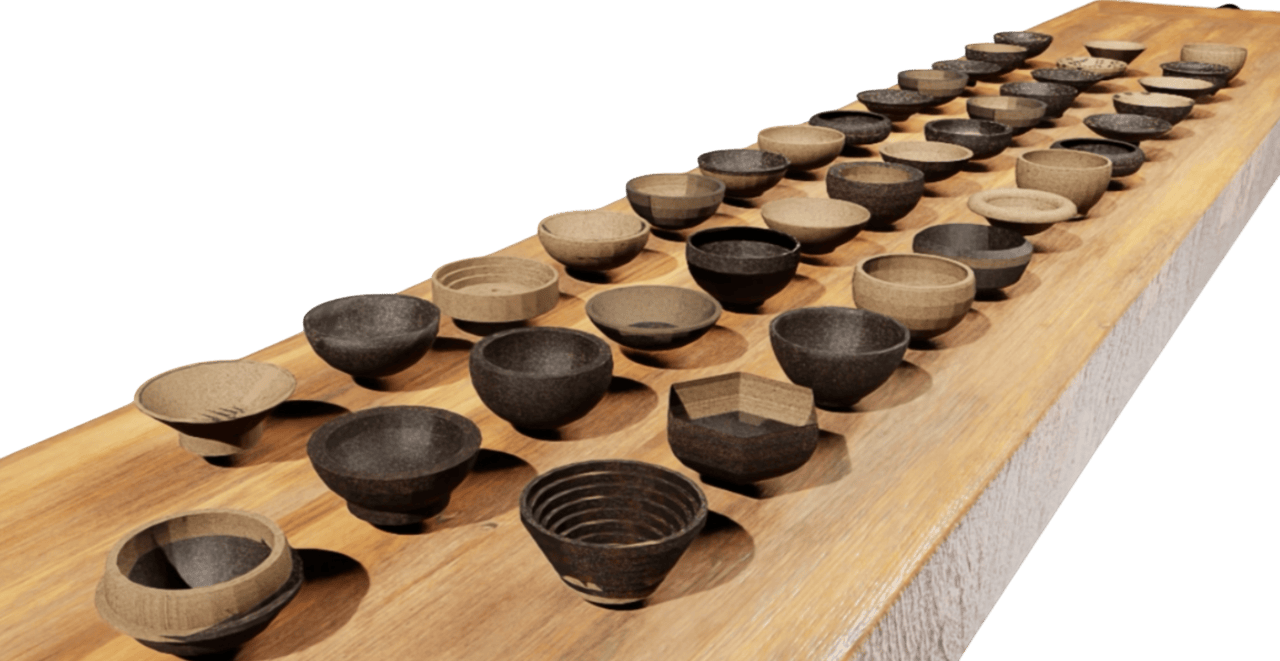}
  \caption{40 bowl variants used in SimVLA, spanning diverse shapes, sizes, depths, and surface materials for robust sim-to-real generalization.}
  \label{fig:bowl_objects}
  \vspace{-1em}
\end{wrapfigure}
\paragraph{Objects.}  To ensure robust generalization across diverse object geometries, SimVLA incorporates a large variety of object instances for each task-relevant category. As shown in Figure~\ref{fig:bowl_objects}, diverse instances for each category (e.g., mugs, bottles, tables, bowls, food items, and utensils) are sourced from Objaverse~\citep{Deitke2023objaverse}, spanning a wide range of shapes, sizes, aspect ratios, and surface materials. During scene generation, objects are sampled with randomized poses, scales, and material properties. This object-level diversity is critical, as policies trained on a narrow set of object instances tend to overfit to specific geometries and fail under even mild shape variation.

\begin{wrapfigure}[11]{r}{0.4\textwidth}
  \vspace{-1em}
  \centering
  \includegraphics[width=\linewidth]{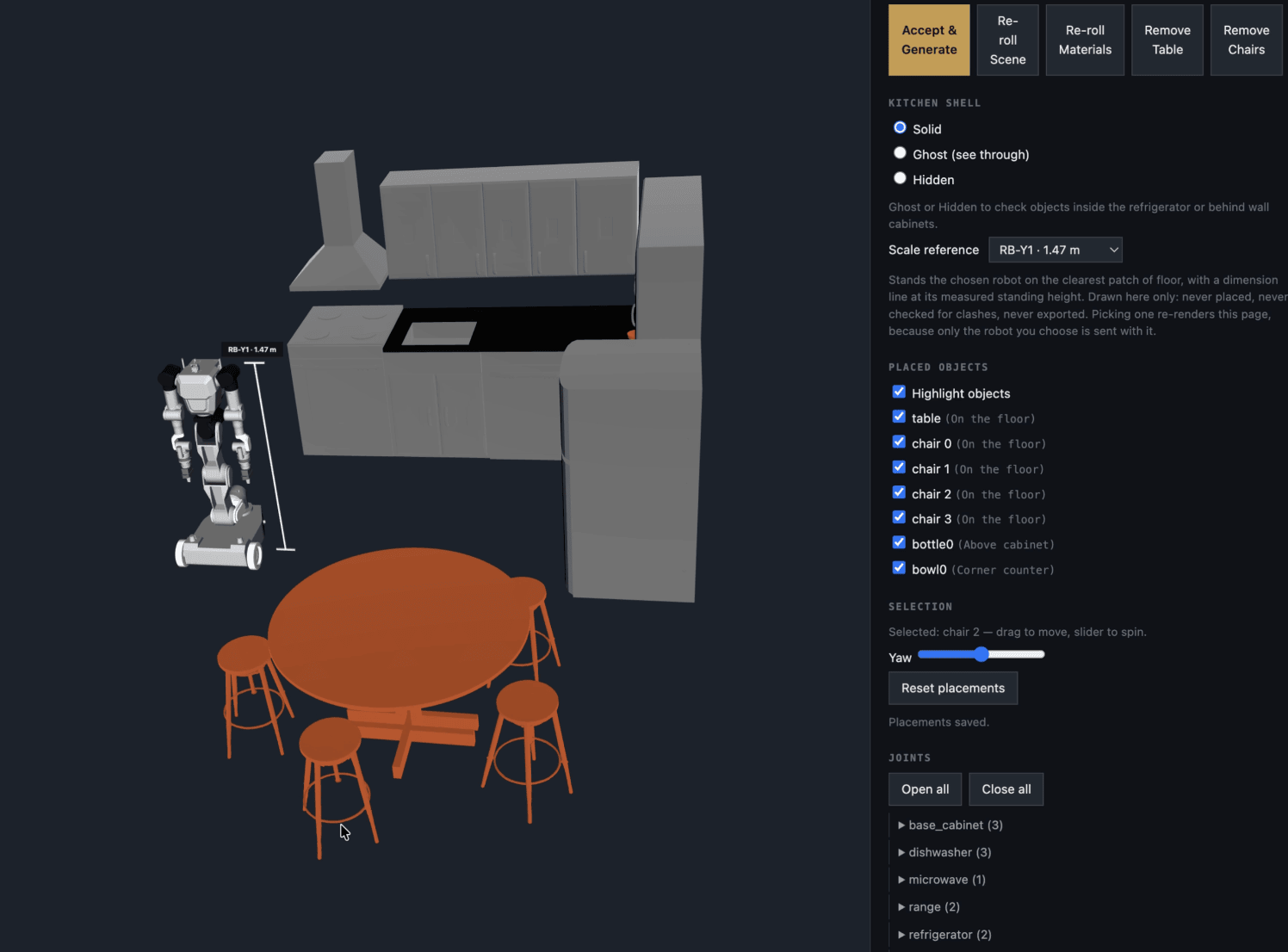}
  \caption{Kitchen scene generation UI.}
  \label{fig:kitchen_scene_ui}
\end{wrapfigure}
\paragraph{Scene Generator.}
To streamline scene construction, SimVLA provides a lightweight GUI for kitchen scene generation, as shown in Figure~\ref{fig:kitchen_scene_ui}. Users specify the kitchen layout type (e.g., L-shaped), select object categories (e.g., bowl, bottle), and assign each object to a semantic placement location (e.g., ``Above cabinet,'' ``Above dishwasher''). The populated object list is displayed before committing, allowing users to review and adjust placements prior to generation. Clicking \textit{Generate Kitchen Scene} instantiates the full scene with randomized object instances, poses, and materials. This interface enables rapid expansion of the training distribution.

\section{Skill API}
\label{sec:skill_api}

\textbf{Atomic Skill Interface.} SimVLA defines a library of atomic skills — including grasping, placing, pouring, opening, closing, navigating, and searching — each implemented as a Python function decorated with \texttt{@skill\_register}. This decorator automatically registers the skill into the task composition framework, making it immediately available for use in long-horizon task definitions. An example of the grasping skill interface is shown in Listing~\ref{lst:skill_grasp}.

\begin{lstlisting}[language=Python,style=srlcode,float=h,
  caption={Atomic grasping skill.},label={lst:skill_grasp}]
@skill_register(name="grasp", arm="right")
def grasp(env, object_name, arm="right"):
    pose = env.grasp_db.sample(object_name, arm)   # Sample from BODex
    move_arm_to_pose(env, arm, pose.intermediate)  # Intermediate grasp
    move_arm_to_pose(env, arm, pose.grasp)         # Final grasp
    close_gripper(env, arm)                        # Close gripper
    return SkillStatus.SUCCESS
\end{lstlisting}

\textbf{Search Skill API.}
Given a target object, the scripted expert uses its ground-truth pose and the
simulator's collision geometry to synthesize a search trajectory in three stages. Ground-truth poses are used only during data generation;
at inference, SimVLA reproduces the learned search behavior from visual observations
alone.
\begin{enumerate}[leftmargin=1.5em,itemsep=1pt,topsep=2pt]
\item \textit{Rotate.} Turn in place until facing a goal position in front of the object.
\item \textit{Approach.} Advance through waypoints toward a standoff point, each
      waypoint laterally perturbed and collision-checked to diversify approach paths.
\item \textit{Align.} Perturb the final heading within the field of view, so the
      object appears at varied image locations rather than always centered.
\end{enumerate}

\begin{figure}[h]
    \includegraphics[width=\linewidth]{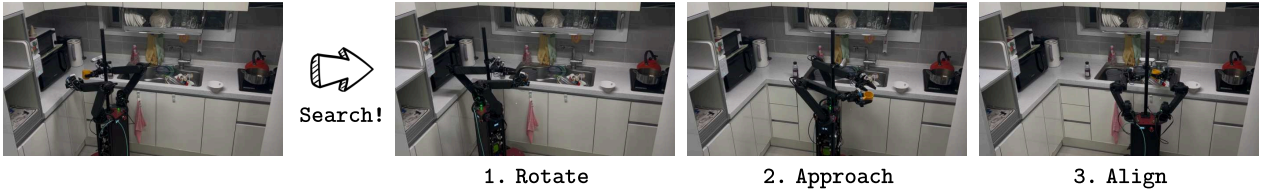}
    \caption{Real-world search behavior. Given only a language prompt and RGB observations, the policy executes searching motions until the sink is visible, then proceeds to complete the task.}
    \label{fig:search_api}
\end{figure}

\begin{wrapfigure}{r}{0.4\textwidth}
    \vspace{-2em}
  \centering
  \includegraphics[width=0.9\linewidth]{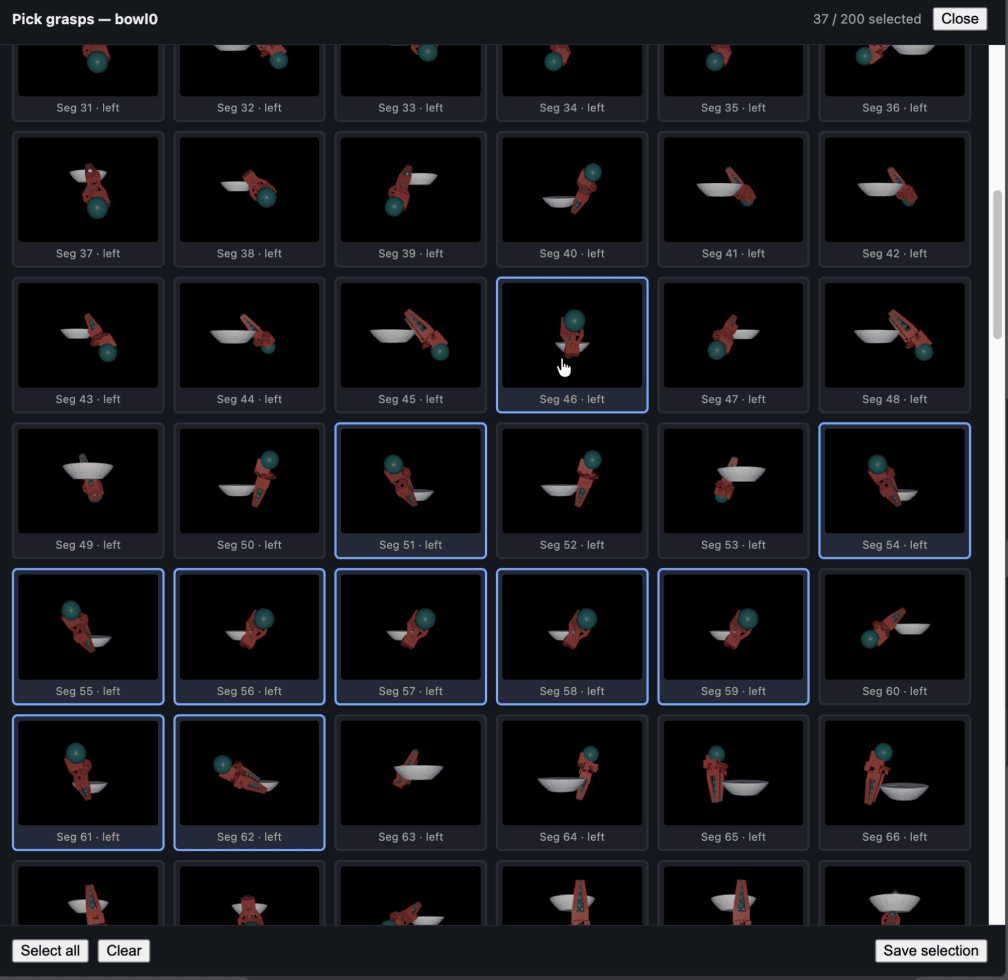}
  \captionof{figure}{BODex grasp pose generation and selection GUI.}
  \label{fig:grasping skill}
  \vspace{-1em}
\end{wrapfigure}

\textbf{Grasp Pose Selection.} For grasping skills, SimVLA integrates BODex~\citep{chen2025bodex} to generate diverse candidate grasp poses for each object. As shown in Figure~\ref{fig:grasping skill}, a GUI presents up to 200 candidates, organized by gripper configuration and arm side (left/right). Users can optionally select preferred grasp poses through the GUI, while the automatically generated candidates provide broad coverage. This human-in-the-loop selection helps ensure that the resulting demonstrations reflect realistic grasping behaviors rather than arbitrary contact configurations.

\clearpage

\section{Hybrid System Identification}
\label{App:Hybrid System Identification}

\begin{wrapfigure}{r}{0.4\textwidth}
    \vspace{-1em}
  \centering
  \includegraphics[width=\linewidth]{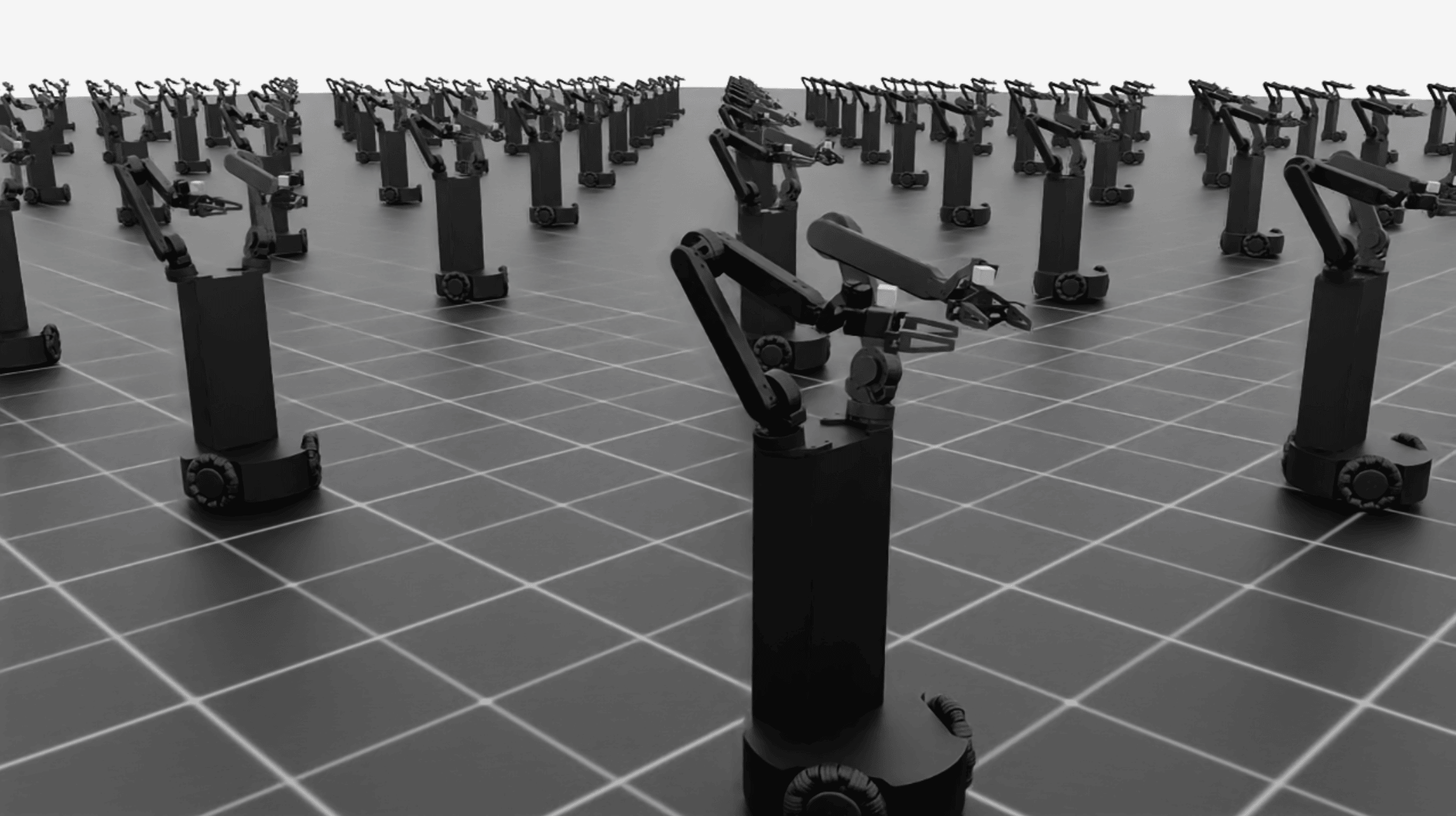}
  \captionof{figure}{Massive parallel environments (4096 instances) running simultaneously in Isaac Lab for hybrid system identification.}
  \label{fig:massive_parallel}
\end{wrapfigure}
Accurate sim-to-real transfer requires that the simulated robot dynamics closely match the real hardware. We perform hybrid system identification to calibrate physical parameters by jointly minimizing end-effector position error, end-effector rotation error, and joint-space MSE between simulation and the real robot. Unlike prior work that matches only joint-space trajectories~\citep{lin2025simtorealrl}, our hybrid objective additionally constrains end-effector tracking, since SimVLA operates in end-effector space. To accelerate the search, we run 4096 parallel simulation environments in Isaac Lab simultaneously, as illustrated in Figure~\ref{fig:massive_parallel}, reducing wall-clock time for parameter calibration from hours to minutes.

\Cref{alg:r2s_autotune_eef} describes the full procedure. A population of $K$ candidate simulation environments is sampled from the parameter space, and each candidate is evaluated against real robot tracking signals collected on $N$ calibration sequences. A candidate is accepted only if all three error metrics — EEF position ($e_p$), EEF rotation ($e_r$), and joint MSE ($e_j$) — remain below their respective thresholds $\tau_p$, $\tau_r$, and $\tau_j$ across all timesteps.

\begin{algorithm}
\caption{Hybrid System Identification}
\label{alg:r2s_autotune_eef}
\begin{algorithmic}[1]
\Require $E$: parameter space; $M$: robot model; $K$: number of parallel
  candidate environments; $N$: number of calibration sequences; $R$: real robot;
  $\tau_p = 0.03$\,m, $\tau_r = 0.2$\,rad, $\tau_j = 0.0025$\,rad$^2$: error thresholds
\Procedure{Autotune}{$E, M, K, N, R, \tau_p, \tau_r, \tau_j$}
    \State $\mathcal{S} \gets \lbrace \Call{SimEnv}{M, p_k} \mid p_k \sim \Call{Sample}{E} \rbrace_{k=1}^{K}$
    \State $\mathcal{U} \gets \Call{GenSeqs}{N}$ \Comment{$N$ calibration sequences}
    \State $\mathbf{R} \gets \Call{GetSignals}{R,\ \mathcal{U}}$ \Comment{Real $\{pos, rot, q\}$}
    \State $J^\star \gets \infty$;\ \ $p^\star \gets \textbf{null}$;\ \ $\mathbf{e}^\star \gets \textbf{null}$
    \For{$k = 1, \dots, K$}
        \State $\mathbf{S} \gets \Call{GetSignals}{S_k,\ \mathcal{U}}$ \Comment{Sim $\{pos, rot, q\}$}
        \State $e_p \gets \max_{n,t}\, \lVert S^{pos}_{n,t} - R^{pos}_{n,t} \rVert_2$ \Comment{EEF position error}
        \State $e_r \gets \max_{n,t}\, \Call{QuatAngleError}{S^{rot}_{n,t}, R^{rot}_{n,t}}$ \Comment{EEF rotation error}
        \State $e_j \gets \mathrm{MSE}(S^{q}, R^{q})$ \Comment{Joint tracking error}
        \State $J \gets e_p/\tau_p + e_r/\tau_r + e_j/\tau_j$ \Comment{Normalized hybrid score}
        \If{$J < J^\star$} \Comment{Keep running minimum}
            \State $J^\star \gets J$;\ \ $p^\star \gets p_k$;\ \ $\mathbf{e}^\star \gets [e_p, e_r, e_j]$
        \EndIf
    \EndFor
    \If{$\mathbf{e}^\star \preceq [\tau_p, \tau_r, \tau_j]$} \Comment{Element-wise threshold check}
        \State \Return $p^\star$
    \EndIf
    \State \Return \textbf{null} \Comment{Best candidate still exceeds a threshold}
\EndProcedure
\end{algorithmic}
\end{algorithm}

\textbf{Ablation study.} 
We compare four calibration settings: no identification (default URDF gains),
joint-space only ($e_j$), EEF-space only ($e_p, e_r$), and our hybrid objective.
For each setting, we replay ten held-out real trajectories spanning three manipulation tasks and measure the EEF position discrepancy between the real and simulated robots, averaged over both arms of the Anubis robot.

\begin{table}[H]
\centering
\caption{System-identification ablation. EEF position discrepancy over ten
held-out trajectories on both arms; $\pm$ denotes standard deviation across trajectories.}
\label{tab:sysid_ablation}
\begin{tabular}{lcc}
\toprule
\textbf{Objective} & \textbf{L2 mean (mm)} $\downarrow$ & \textbf{L2 max (mm)} $\downarrow$ \\
\midrule
None (URDF defaults) & $83.4 \pm 50.2$ & $120.7 \pm 96.0$ \\
Joint-space          & $14.7 \pm 2.3$  & $46.2 \pm 9.6$ \\
EEF-space            & $13.4 \pm 1.7$  & $40.8 \pm 8.4$ \\
Hybrid (joint + EEF) & $\mathbf{13.3 \pm 1.6}$ & $\mathbf{40.6 \pm 8.1}$ \\
\bottomrule
\end{tabular}
\end{table}

Calibration has a substantial effect: hybrid identification cuts mean EEF error by
$6.3\times$ ($83.4 \to 13.3$\,mm) and maximum error by $3.0\times$
($120.7 \to 40.6$\,mm) over default gains, while sharply reducing variance across
trajectories. The choice of objective has a much smaller effect. Hybrid is best on both metrics but gains only $1.4$\,mm in mean error over joint-space identification and is within noise of EEF-space identification, suggesting that the primary benefit comes from performing system identification itself.

\newpage

\section{SimAction}
\label{App:SimVLA Action Data}
SimAction comprises long-horizon manipulation tasks and atomic manipulation skills in kitchen environments. These tasks involve navigation, object grasping, articulated manipulation, object placement, and sequential task execution. SimVLA includes 35 tasks, comprising 22 long-horizon compositions and 13 atomic skills, with each atomic skill also released as a standalone task. Each task provides three RGB streams (front, left wrist, and right wrist) at $240{\times}320$ resolution, recorded at 20 fps. The robot action space is 23-D, comprising a 10-D end-effector representation for each arm (3-D position, 6-D rotation and 1-D gripper), together with the base velocities $(v_x, v_y, \omega)$. We additionally provide the proprioceptive state and the initial robot pose. The SimAction tasks are listed below.

\begin{table*}[h]
\centering
\fontsize{8pt}{9pt}\selectfont
\setlength{\tabcolsep}{4pt}
\caption{Task definitions.}
\label{tab:tasks}
\begin{tabular}{@{}r l p{0.70\textwidth}@{}}
\toprule
ID & Task & Description \\
\midrule
1  & Move to mug & The robot navigates and positions itself near the mug. \\
2  & Move to bowl & The robot navigates and positions itself near the bowl. \\
3  & Move to bottle & The robot navigates and positions itself near the bottle. \\
4  & Move to drawer & The robot navigates and positions itself in front of the drawer. \\
5  & Put mug in sink & The robot navigates to the mug, grasps it, moves to the sink, and places it into the sink. \\
6  & Put bowl in sink & The robot navigates to the bowl, grasps it, moves to the sink, and places it into the sink. \\
7  & Put bottle in sink & The robot navigates to the bottle, grasps it, moves to the sink, and places it into the sink. \\
8  & Put mug on table & The robot navigates to the mug, grasps it, moves to the table, and gently places it on the table. \\
9  & Put bowl on table & The robot navigates to the bowl, grasps it, moves to the table, and gently places it on the table. \\
10 & Put bottle on table & The robot navigates to the bottle, grasps it, moves to the table, and gently places it on the table. \\
11 & Put apple on table & The robot navigates to the apple, grasps it, moves to the table, and gently places it on the table. \\
12 & Put apple in bowl & The robot navigates to and grasps the apple, moves to the bowl, and places the apple into the bowl. \\
13 & Put apple beside bowl & The robot navigates to and grasps the apple, moves to the bowl, and places the apple beside the bowl. \\
14 & Put apple beside mug & The robot navigates to and grasps the apple, moves to the mug, and places the apple beside the mug. \\
15 & Serve bowl & The robot navigates to and grasps a bowl, transports it to a serving location, and places it down. \\
16 & Serve mug & The robot navigates to and grasps a mug, transports it to a serving location, and places it down. \\
17 & Put pot on stove & The robot navigates to the pot, grasps it, moves to the stove, and places it onto the stove. \\
18 & Put ramen in pot & The robot grasps ramen ingredients and places them into the pot. \\
19 & Place bowl in drawer & The robot places a grasped bowl into the opened drawer. \\
20 & Open drawer & The robot grasps the drawer handle and opens the drawer. \\
21 & Close drawer & The robot closes the drawer after manipulation. \\
22 & Open microwave & The robot moves to the microwave, grasps the handle, and opens it by moving its arm. \\
23 & Close microwave & The robot moves to the microwave, grasps the handle, and closes it by moving its arm. \\
24 & Open refrigerator & The robot navigates to the refrigerator, grasps the handle, and opens the refrigerator door. \\
25 & Close refrigerator & The robot grasps the handle and closes the refrigerator door. \\
26 & Open dishwasher & The robot moves to the dishwasher, grasps the handle, and opens it by moving its arm and base. \\
27 & Close dishwasher & The robot moves to the dishwasher, grasps the handle, and closes it by moving its arm and base. \\
28 & Pour water into pot & The robot grasps a water container and pours water into the pot. \\
29 & Push chair & The robot navigates to the chair and pushes it toward the table using both arms. \\
30 & Put mug and bowl in sink & The robot sequentially navigates to and grasps both the mug and the bowl, moves to the sink, and places them into the sink. \\
31 & Put bowl in drawer & The robot navigates to and grasps the bowl, moves to the drawer, opens the drawer, places the bowl inside, and closes the drawer. \\
32 & Put mug in drawer & The robot navigates to and grasps the mug, moves to the drawer, opens the drawer, places the mug inside, and closes the drawer. \\
33 & Pour from bottle to mug & The robot navigates to and grasps the bottle and the mug, then pours from the bottle into the mug. \\
34 & Boil sweet potato & The robot places sweet potatoes into a pot and performs the cooking procedure. \\
35 & Prepare snack & The robot sequentially manipulates multiple food-related objects to prepare a snack setup. \\
\bottomrule
\end{tabular}
\end{table*}

\newpage

\section{Domain Randomization}
\label{Domain Randomization}

To bridge the sim-to-real gap, we apply extensive domain randomization across six categories: camera perturbations, kitchen scene properties, object physical and visual properties, robot dynamics, and lighting conditions. Rather than relying solely on system identification to match real-world dynamics, domain randomization exposes the policy to a broad distribution of physical and visual conditions during training, encouraging it to learn behaviors that are robust to the inevitable discrepancies between simulation and reality. The full randomization setup is summarized in Table~\ref{tab:Domain Randomization Setup}.

\begin{table*}[h]
\centering
\footnotesize % smaller
\setlength{\tabcolsep}{4pt}
\renewcommand{\arraystretch}{1.0}
\caption{Domain Randomization Setup.}
\label{tab:Domain Randomization Setup}
\begin{tabularx}{\textwidth}{Xc}
\toprule
\textbf{Domain} & \textbf{Randomness} \\
\midrule
Camera: Position Noise (m) & $\sim \mathcal{N}(0, 0.03^{2})$ \\
Camera: Rotation Noise (deg) & $\sim \mathcal{U}(-5, 5)$ \\
\midrule
Kitchen: Types & 5 \\
Kitchen: Tables & 50 \\
Kitchen: Chairs & 31 \\
Kitchen: Cabinet Materials & 21 \\
Kitchen: Wall Materials & 33 \\
Kitchen: Floor Materials & 18 \\
Kitchen: Cabinet Height (m) & $\sim \mathcal{N}(0.85, 0.03^{2})$ \\
Kitchen: Articulation Furniture PD Controller: P Gain & $\sim \mathcal{U}(1, 25)$ \\
Kitchen: Articulation Furniture PD Controller: D Gain & $\sim \mathcal{U}(0, 1)$ \\
Kitchen: Handle Static Contact Friction Coefficient & $\sim \mathcal{U}(0.3, 0.8)$ \\
Kitchen: Handle Dynamic Contact Friction Coefficient & $\sim \mathcal{U}(0.2, 0.6)$ \\
Kitchen: Handle Position & $\sim \mathcal{U}(-0.05, 0.05)$ \\
\midrule
Object: Mass (kg) & $\sim \mathcal{U}(0.1, 0.5)$ \\
Object: Static Contact Friction Coefficient & $\sim \mathcal{U}(0.2, 0.6)$ \\
Object: Dynamic Contact Friction Coefficient & $\sim \mathcal{U}(0.1, 0.5)$ \\
Object: Restitution Coefficient & $\sim \mathcal{U}(0.3, 0.7)$ \\
Object: Shape & $\times \sim \mathcal{N}(1, 0.07^{2})$ \\
Object: Initial Position Noise (m) & $\sim \mathcal{U}(-0.1, 0.1)$ \\
Object: Initial z-orientation (deg) & $\sim \mathrm{Cat}(\{0, 30, 60,\ldots,330\})$ \\
Object: Center of Mass (x-axis) (m) & $\sim \mathcal{U}(-0.005, 0.005)$ \\
Object: Center of Mass (y-axis) (m) & $\sim \mathcal{U}(-0.005, 0.005)$ \\
Object: Center of Mass (z-axis) (m) & $\sim \mathcal{U}(-0.02, 0.02)$ \\
Object: Color (RGB) & $\sim \mathcal{U}(0.0, 1.0)$ \\
Object: Mug Types & 80 \\
Object: Bowl Types & 40 \\
Object: Bottle Types & 20 \\
\midrule
Robot: Initial Position & Anywhere in the room without collision \\
Robot: Initial Rotation (deg) & $\sim \mathcal{U}(0, 360)$ \\
Robot: Initial Joint Position Noise & $\sim \mathcal{U}(-0.1, 0.1)$ \\
Robot: Initial Joint Velocity Noise & $\sim \mathcal{U}(-0.01, 0.01)$ \\
Robot: Static Contact Friction Coefficient & $\sim \mathcal{U}(0.7, 1.0)$ \\
Robot: Dynamic Contact Friction Coefficient & $\sim \mathcal{U}(0.5, 1.0)$ \\
Robot: Restitution Coefficient & $\sim \mathcal{U}(0.3, 0.5)$ \\
Robot: PD Controller (P Gain) & $\times \sim \mathcal{N}(1, 0.1^{2})$ \\
Robot: PD Controller (D Gain) & $\times \sim \mathcal{N}(1, 0.1^{2})$ \\
Robot: Color (RGB range [0,1]) & $\sim \mathcal{U}(0.0, 0.1)$ \\
\midrule
Light: Intensity scale & $\sim 2^{\mathcal{U}(-1.5,\, 1.5)}\ (I = 3000 \cdot \text{scale})$ \\
Light: Color temperature (K) & $\sim \mathrm{Cat}\{2700, 3200, 4000, 5000, 6500, 8000, 9000\}$ \\
Light: Color (RGB) & $\sim \mathrm{clip}\!\left(\mathbf{1} + \mathcal{N}(0, 0.08^{2}),\, [0.7, 1.3]\right)$ \\
\bottomrule
\end{tabularx}
\end{table*}

\newpage

\section{SimVQA}
\label{App:SimVQA}

SimVQA is a visual question-answering dataset automatically generated from simulation to inject geometric and subtask-level supervision into the VLM backbone. By leveraging ground-truth simulator state, SimVQA provides precise spatial labels that would be prohibitively expensive to obtain from real-world data. Table~\ref{tab:vqa_prompt_templates} lists the full set of VQA templates used, covering high-level task planning, subtask sequencing, object detection, 3D spatial reasoning, cross-view correspondence, reachability estimation, graspability assessment, and gripper state queries. Our dataset contains approximately 1M SimVQA visual question-answering samples.

\newcommand{\bk}{\discretionary{}{}{}}          % zero-width, hyphen-free break
\newcommand{\func}[1]{\texttt{\small #1}}

\begin{table}[H]
\centering
\caption{SimVQA Templates}
\label{tab:vqa_prompt_templates}
\renewcommand{\arraystretch}{1.15}
\setlength{\tabcolsep}{6pt}
\begin{tabularx}{\linewidth}{
  >{\raggedright\arraybackslash}p{0.27\linewidth}
  >{\raggedright\arraybackslash}X
  >{\raggedright\arraybackslash}p{0.22\linewidth}
}
\toprule
\textbf{VQA Function} & \textbf{Question Prompt Prototype} & \textbf{Answer Prototype} \\
\midrule
\func{high\_level\_\bk subtask} &
To complete the task ``\{task\}'', what should the robot do now? &
\{subtask\} \\
\func{next\_subtask} &
After current subtask \{subtask\}, what is the next subtask? &
\{subtask\} \\
\func{object\_detection} &
Is the \{object\} visible in the front camera? &
yes. bbox: \{y1,x1,y2,x2\} / no \\
\func{object\_3d\_\bk information} &
How far is the \{object\} from the robot base? &
dx \{x\} dy \{y\} dz \{z\} \\
\func{view\_\bk correspondence} &
Given front bbox \{bbox\}, what are the wrist camera bboxes? &
left \{bbox\} right \{bbox\} \\
\func{object\_reachable} &
Is the robot close enough to manipulate the \{object\}? &
yes / no. move base dx \{x\} dy \{y\} dyaw \{yaw\} \\
\func{object\_graspable} &
Is the \{object\} graspable with left gripper? &
yes / no. move gripper dx \{x\} dy \{y\} dz \{z\} dquat \{quat\} \\
\func{left\_gripper\_open} &
Is the robot's left gripper closed? &
yes / no \\
\func{right\_gripper\_open} &
Is the robot's right gripper closed? &
yes / no \\
\bottomrule
\end{tabularx}
\end{table}

\newpage

\section{SimVLA Training}
\label{sec:training_details}

SimVLA follows a two-stage training pipeline built on a pre-trained PaliGemma VLM backbone and a flow-matching action expert. The two stages differ in which losses are active and whether gradients are allowed to flow back into the VLM backbone.

\paragraph{Pre-training: Knowledge-isolated co-training.}
\label{sec:simvla_pretraining}
We denote the input tokens of an example by $x = (x^{p}, x^{f}, x^{c}, x^{a})$, where $x^{p}$ are the \emph{prefix} tokens (image patches encoded by SigLIP, language instruction, and proprioceptive state), $x^{f}$ are FAST-tokenized action tokens~\citep{pertsch2025fast}, $x^{c}$ are SimVQA language tokens, and $x^{a}=a^{\tau,\omega}_{1:H}$ are noisy continuous-action tokens with $a^{\tau,\omega}_{i}=\tau a_{i}+(1-\tau)\omega_{i}$, where $\tau\sim p(\tau)$ and $\omega\sim\mathcal{N}(0,I)$. The flow-matching action expert $f^{a}_{\theta}$ predicts the vector field from the noisy action $x^a$ toward the clean action trajectory, conditioned on the prefix and interpolation time $\tau$. For SimVLA, we use $H{=}50$ and $d{=}23$.

Each example $b$ is assigned three Bernoulli masks $M^{a}_{b},M^{f}_{b},M^{c}_{b}\in\{0,1\}$, which determine whether the flow-matching, FAST, and VQA losses, respectively, are applied to that example. Our work follows the knowledge insulation framework~\citep{driess2025knowledge}. The three objectives are optimized in a single forward pass:
\begin{align}
    \mathcal{L}_{\text{flow}}(\theta)
    &=
    \mathbb{E}_{\mathcal{D},\tau,\omega}\Big[
        M^{a}\,
        \big\|\,
        (a_{1:H}-\omega)
        -
        f^{a}_{\theta}\!\left(x^{a};\,x^{p},\tau\right)
        \big\|^{2}
    \Big],
    \label{eq:simvla-flow}\\
    \mathcal{L}_{\text{fast}}(\theta)
    &=
    \mathbb{E}_{\mathcal{D}}\Big[
        -M^{f}
        \sum_{j=1}^{n_f}
        \log p_{\theta}\!\left(
            x^{f}_{j}\mid x^{p},x^{f}_{<j}
        \right)
    \Big],
    \label{eq:simvla-fast}\\
    \mathcal{L}_{\text{vqa}}(\theta)
    &=
    \mathbb{E}_{\mathcal{D}}\Big[
        -M^{c}
        \sum_{j=1}^{n_c}
        \log p_{\theta}\!\left(
            x^{c}_{j}\mid x^{p},x^{c}_{<j}
        \right)
    \Big],
    \label{eq:simvla-vqa}\\
    \mathcal{L}_{\text{pre}}(\theta)
    &=
        \mathcal{L}_{\text{flow}}(\theta)
        +\alpha_f\,\mathcal{L}_{\text{fast}}(\theta)
        +\alpha_c\,\mathcal{L}_{\text{vqa}}(\theta).
    \label{eq:simvla-pre}
\end{align}
Here, $n_f$ and $n_c$ denote the lengths of the FAST-action and SimVQA token sequences, respectively, and $x^f_{<1}=x^c_{<1}=\emptyset$. Thus, the first FAST or VQA token is predicted directly from the prefix $x^p$, while subsequent tokens are predicted autoregressively from the preceding postfix tokens.

To prevent the randomly initialized action expert from corrupting the pre-trained VLM backbone, we apply knowledge insulation. Specifically, gradients from the flow loss are stopped at the VLM backbone parameters $\theta_b$, while the FAST and VQA losses update the backbone normally. The attention mask is additionally modified to prevent cross-attention between continuous action tokens and FAST/VQA tokens, ensuring that the three supervision streams remain decoupled within the same forward pass.

\paragraph{Post-training.}
Once the action expert is warmed up during pre-training, we lift the stop-gradient so that the flow loss can back-propagate into the VLM backbone $\theta_b$. The FAST and VQA losses are dropped at this stage, and the model is fine-tuned exclusively on mobile-manipulation tasks matching the downstream evaluation setting. The post-training objective is
\begin{align}
    \mathcal{L}_{\text{post}}(\theta_b,\theta_a)
    &=
    \mathbb{E}_{\mathcal{D}_{\text{post}},\tau,\omega}\Big[
        \big\|
        (a_{1:H}-\omega)
        -
        f^{a}_{\theta_b,\theta_a}
        \!\left(x^{a};\,x^{p},\tau\right)
        \big\|^{2}
    \Big],
    \label{eq:simvla-post}
\end{align}
where $\theta_a$ and $\theta_b$ denote the action-expert and VLM backbone parameters, respectively, and gradients from the flow loss are allowed to update both parameter sets.

We train on
$\mathcal{D}_{\text{post}}
=
\mathcal{D}_{\text{cont}}\cup\mathcal{D}_{\text{deploy}}$,
where $\mathcal{D}_{\text{cont}}$ is the original continuous-action corpus used in pre-training and $\mathcal{D}_{\text{deploy}}$ is the mobile-manipulation deployment dataset collected from policy rollouts across diverse simulated environments and matched to the downstream evaluation setting. Because the action expert is already warmed up, allowing gradients to flow into the backbone during this stage enables task-specific adaptation while avoiding the instability that could arise from jointly training a randomly initialized action expert with the pre-trained VLM representations.

\paragraph{Hyperparameters and Compute.}
SimVLA is trained in two stages using AdamW with cosine learning rate scheduling. Tables~\ref{tab:simvla-hparams}--\ref{tab:simvla-compute} summarize the full hyperparameter configuration and compute resources used for pre-training and post-training. In total, the reported experiments required approximately 1{,}272 GPU-hours.
Beyond the experiments included in the paper, additional compute was used for
preliminary experiments, hyperparameter tuning, failed runs, and debugging.
These additional runs required substantially more compute than the final reported
experiments.

% ---------- Table 1: pretraining + post-training hyperparameters ------
\begin{table}[H]
\centering
\caption{Key hyperparameters for SimVLA training.}
\label{tab:simvla-hparams}
\begin{tabular}{lcc}
\toprule
Hyperparameter & Pretrain & Post-train \\
\midrule
Global batch size           & 256                & 256                \\
Precision                   & BF16               & BF16               \\
Gradient clipping (L2)      & 1.0                & 1.0                \\
Peak learning rate          & $5\times10^{-5}$   & $5\times10^{-5}$   \\
End learning rate           & $5\times10^{-5}$   & $5\times10^{-5}$   \\
LR scheduler                & cosine             & cosine             \\
Warm-up steps               & 1{,}000                  & 1{,}000            \\
Warm-up ratio               & 0.00               & 0.01               \\
Decay steps                 & 180k               & 50k               \\
Total steps                 & 180k               & 50k               \\
Optimizer                   & AdamW              & AdamW              \\
Adam $\beta_1, \beta_2$     & 0.9, 0.95          & 0.9, 0.95          \\
Adam $\epsilon$             & $1\times10^{-8}$   & $1\times10^{-8}$   \\
Weight decay                & $1\times10^{-10}$  & $1\times10^{-10}$  \\
EMA decay                   & 0.999              & 0.999              \\
Vision backbone frozen      & false              & false              \\
Action chunk size           & 50                 & 50                 \\
Sampling steps              & 10                 & 10                 \\
Data workers                & 32                 & 32                 \\
Weight init                 & Paligemma base   & SimVLA pretrain ckpt \\
\bottomrule
\end{tabular}
\end{table}

% ---------- Table 2: co-training / FAST auxiliary hyperparameters -----
\begin{table}[H]
\centering
\caption{Co-training and FAST auxiliary-loss hyperparameters used during SimVLA pretraining.}
\label{tab:simvla-cotrain}
\begin{tabular}{lc}
\toprule
Hyperparameter & Value \\
\midrule
FAST loss weight           & 0.1   \\
VQA loss weight       & 0.1   \\
FAST sampling ratio        & 0.50  \\
Co-train sampling ratio    & 0.01  \\
Discrete state input       & false \\
Delta joint actions        & true  \\
\bottomrule
\end{tabular}
\end{table}

% ---------- Table 3: compute resources --------------------------------
\begin{table}[H]
\centering
\caption{Compute resources used for SimVLA experiments. All experiments were
conducted on an internal GPU cluster. Reported times are approximate wall-clock
durations and may vary depending on storage and data-loading throughput.}
\label{tab:simvla-compute}
\resizebox{\textwidth}{!}{%
\begin{tabular}{lcccccc}
\toprule
Stage & Hardware & CPU & System RAM & Storage & Per-run compute & Total compute \\
\midrule
Pre-training
& 4$\times$ NVIDIA B200 (183GB)
& 128 CPU cores
& 1.5 TB
& NVMe SSD
& $\sim$12 days
& $\sim$1152 GPU-hours \\
Post-training
& 4$\times$ NVIDIA B200 (183GB)
& 128 CPU cores
& 1.5 TB
& NVMe SSD
& $\sim$30 hours
& $\sim$120 GPU-hours \\
\bottomrule
\end{tabular}%
}
\end{table}

\newpage

\section{Experimental Setup}
\label{App:Experimental Setup}

\paragraph{Robot Hardware Spec}
We conduct our real-world experiments using Anubis robot~\citep{kang2025anubis}. The platform features a teleoperation system inspired by the Mobile ALOHA setup~\citep{fu2024mobilealoha}. Each arm has 6 DoF and is equipped with a parallel gripper and a wrist-mounted camera. At the center of the robot, an Intel RealSense camera is mounted on a height-adjustable mechanism, serving as the ego-centric view camera. Details are described in~\Cref{tab:anubis_hardware}. Anubis is equipped with a 3-wheel omni-directional base that supports planar locomotion.

\begin{table}[h]
  \centering
  \begin{minipage}[b]{0.65\textwidth}
    \centering
    \footnotesize
    \setlength{\tabcolsep}{4pt}
    \vspace{1em}
    \caption{\textbf{Anubis Robot Hardware Specifications.}}
    \begin{tabularx}{\linewidth}{@{}lX@{}}
      \toprule
      \textbf{Component}    & \textbf{Specification} \\
      \midrule
      Base Type             & 3-wheel omni-directional chassis \\
      Mobility DOF          & 3 (X, Y, Yaw) \\
      Arm DOF               & 2 × (6 DOF + gripper) = 14 \\
      Total DOF             & 14 arm + 3 base = 17 \\
      Policy action space   & 2 × (9D EEF + 1D gripper) + 3D base = 23-D\\
      Front Camera          & Intel RealSense D455 \\
      Wrist Cameras         & Intel RealSense D405 \\
      Gripper               & Parallel transparent gripper (hole design, ALOHA-style) \\
      Power System          & 3 × Greenworks 40V 5.0Ah batteries (PC, wheels \& leader/follower) \\
      Frame                 & 3D-printed custom components \\
      \bottomrule
    \end{tabularx}
    \label{tab:anubis_hardware}
  \end{minipage}
  \hfill
  \begin{minipage}[b]{0.30\textwidth}
    \centering
    \includegraphics[width=0.6\linewidth]{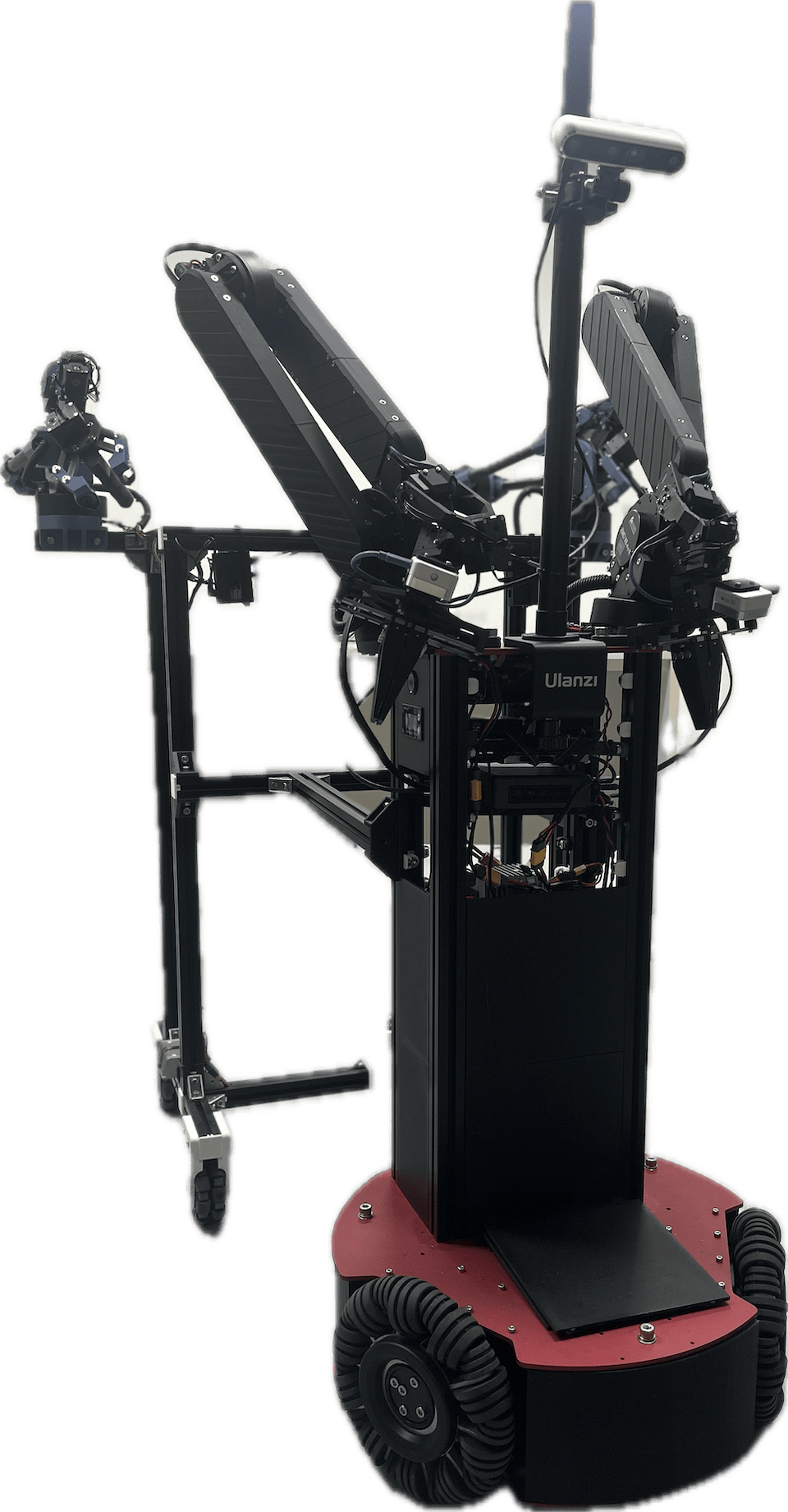} 
    \vspace{1em}
    \captionof{figure}{\textbf{The Anubis robot.}}
    \label{fig:anubis}
  \end{minipage}
\end{table}

\paragraph{Task Progress Description.}
\label{Task progress description}
We define discrete task progress labels for each task to provide finer-grained evaluation beyond binary success metrics.

\begin{itemize}
    \item \textbf{Task 1: Put mug in sink}
    \begin{itemize}
        \item \textbf{0}: Failed grasp.
        \item \textbf{1}: Failed to place the mug into the sink.
        \item \textbf{2}: Success.
    \end{itemize}

    \item \textbf{Task 2: Put bowl in drawer}
    \begin{itemize}
        \item \textbf{0}: Failed grasp.
        \item \textbf{1}: Failed to grasp the drawer handle.
        \item \textbf{2}: Failed to open the drawer (e.g., slipped during opening).
        \item \textbf{3}: Failed to place the bowl into the drawer.
        \item \textbf{4}: Failed to close the drawer.
        \item \textbf{5}: Success.
    \end{itemize}

    \item \textbf{Task 3: Pour from bottle to mug}
    \begin{itemize}
        \item \textbf{0}: Failed to grasp the bottle or mug.
        \item \textbf{1}: Failed to grasp both objects.
        \item \textbf{2}: Failed to tilt the bottle for pouring.
        \item \textbf{3}: Success.
    \end{itemize}
    
    \item \textbf{Task 4: Put bowl in sink}
    \begin{itemize}
        \item \textbf{0}: Failed grasp.
        \item \textbf{1}: Failed to place the bowl into the sink.
        \item \textbf{2}: Success.
    \end{itemize}
    
    \item \textbf{Task 5: Put bottle in sink}
    \begin{itemize}
        \item \textbf{0}: Failed grasp.
        \item \textbf{1}: Failed to place the bottle into the sink.
        \item \textbf{2}: Success.
    \end{itemize}
\end{itemize}

\begin{table*}[h]
\centering
\caption{\textbf{Held-out 1,369 SimVQA evaluation.}
Ablation study on visual reasoning capabilities. $\uparrow$ indicates
higher is better and $\downarrow$ indicates lower is better.}
\label{tab:ablation_visual_reasoning}

\resizebox{0.92\textwidth}{!}{%
\begin{tabular}{lccccc}
\toprule
\textbf{Model} &
\textbf{Gripper} $\uparrow$ &
\textbf{Subtask} $\uparrow$ &
\textbf{3D Pos.} $\downarrow$ &
\textbf{Grasp} $\downarrow$ &
\textbf{Cross-view IoU} $\uparrow$ \\
& \textbf{L / R} &
\textbf{Curr. / Next} &
\textbf{(m)} &
\textbf{(m)} &
\\
\midrule
SimVLA w/o SimVQA
& --$^{*}$ & --$^{*}$ & --$^{*}$ & --$^{*}$ & --$^{*}$ \\
SimVLA w/o TokSimAction
& 0.86 / 0.69 & 0.18 / 0.39 & 0.50 & 0.16 & 0.04 \\
SimVLA w/o both
& 0.85 / 0.68 & --$^{\dagger}$ & --$^{\dagger}$ & --$^{\dagger}$ & --$^{\dagger}$ \\
\midrule
\textbf{SimVLA}
& \textbf{1.00 / 0.89}
& \textbf{0.69 / 0.77}
& \textbf{0.06}
& \textbf{0.05}
& \textbf{0.61} \\
\bottomrule
\end{tabular}%
}

\vspace{3pt}

\begin{minipage}{0.92\textwidth}
\scriptsize
$^{*}$ Emits tokenized actions; metrics are undefined.
\hspace{1em}
$^{\dagger}$ Emits invalid answers; metrics are undefined.
\end{minipage}

\end{table*}

\paragraph{Long-Horizon Simulation Evaluation.}
To provide evidence beyond the real-world tasks, we post-trained and evaluated SimVLA on eight long-horizon tasks in simulation using Anubis robot. We report the success rate over 50 rollouts per task across five kitchen setups (10 rollouts per setup). The results are summarized in Table~\ref{tab:long_horizon_sim}. These tasks cover several distinct skill compositions, including visual navigation, grasping different object types (mug, bowl, and bottle), placement into sinks and drawers, articulated drawer opening and closing, and bimanual pouring. The lower success rate for \textit{Put mug and bowl to sink} (19/50) is expected, as the task chains two complete navigation and pick-and-place sequences, increasing the number of opportunities for failure.

\begin{table}[h]
    \centering
    \caption{SimVLA success rates on eight long-horizon simulation tasks using Anubis robot. 
    Each task is evaluated over 50 rollouts across five unseen kitchen setups.}
    \label{tab:long_horizon_sim}
    \begin{tabular}{lc}
        \toprule
        \textbf{Task} & \textbf{Success Rate} \\
        \midrule
        Put mug to sink & 76\% \\
        Put bowl in drawer & 70\% \\
        Pour from bottle to mug & 76\% \\
        Put mug and bowl to sink & 38\% \\
        Put bowl to sink & 62\% \\
        Put bottle to sink & 64\% \\
        Put mug in drawer & 44\% \\
        Put bottle in drawer & 58\% \\
        \bottomrule
    \end{tabular}
\end{table}

\begin{table}[t]
\centering\small
\renewcommand{\arraystretch}{1.15}
\setlength{\tabcolsep}{4pt}
\caption{\textbf{Mock Kitchen --- ID.} The $\pi_{0.5}$ model fine-tuned with 200 demonstrations achieves the best performance on all tasks. SimVLA outperforms $\pi_{0.5}$ fine-tuned on 50 real-world demonstrations in task progress across all tasks, with comparable or higher binary success. Task progress $\pm$ s.d.\ over 20 trials; binary successes in parentheses.}
\label{tab:mock_id}
\begin{tabular}{lccccc}
\toprule
Model & Task 1 & Task 2 & Task 3 & Task 4 & Task 5 \\
\midrule
$\pi_{0.5}$ fine-tuned w/ 50 demos & \shortstack{0.55\,$\pm$\,0.43\\[1pt] {\scriptsize (8/20)}} & \shortstack{0.36\,$\pm$\,0.39\\[1pt] {\scriptsize (5/20)}} & \shortstack{0.48\,$\pm$\,0.38\\[1pt] {\scriptsize (6/20)}} & \shortstack{0.45\,$\pm$\,0.46\\[1pt] {\scriptsize (7/20)}} & \shortstack{0.43\,$\pm$\,0.47\\[1pt] {\scriptsize (7/20)}} \\
$\pi_{0.5}$ fine-tuned w/ 200 demos & \shortstack{\textbf{0.75\,$\pm$\,0.44}\\[1pt] {\scriptsize (15/20)}} & \shortstack{\textbf{0.66\,$\pm$\,0.42}\\[1pt] {\scriptsize (11/20)}} & \shortstack{\textbf{0.70\,$\pm$\,0.36}\\[1pt] {\scriptsize (10/20)}} & \shortstack{\textbf{0.65\,$\pm$\,0.43}\\[1pt] {\scriptsize (11/20)}} & \shortstack{\textbf{0.63\,$\pm$\,0.46}\\[1pt] {\scriptsize (11/20)}} \\
SimVLA (w/o pre-train) & \shortstack{0.28\,$\pm$\,0.38\\[1pt] {\scriptsize (3/20)}} & \shortstack{0.15\,$\pm$\,0.16\\[1pt] {\scriptsize (0/20)}} & \shortstack{0.15\,$\pm$\,0.25\\[1pt] {\scriptsize (1/20)}} & \shortstack{0.18\,$\pm$\,0.34\\[1pt] {\scriptsize (2/20)}} & \shortstack{0.18\,$\pm$\,0.29\\[1pt] {\scriptsize (1/20)}} \\
SimVLA (w/o SimDeploy) & \shortstack{0.45\,$\pm$\,0.32\\[1pt] {\scriptsize (3/20)}} & \shortstack{0.38\,$\pm$\,0.31\\[1pt] {\scriptsize (3/20)}} & \shortstack{0.45\,$\pm$\,0.33\\[1pt] {\scriptsize (4/20)}} & \shortstack{0.33\,$\pm$\,0.34\\[1pt] {\scriptsize (2/20)}} & \shortstack{0.33\,$\pm$\,0.34\\[1pt] {\scriptsize (2/20)}} \\
SimVLA & \shortstack{0.60\,$\pm$\,0.42\\[1pt] {\scriptsize (9/20)}} & \shortstack{0.50\,$\pm$\,0.36\\[1pt] {\scriptsize (6/20)}} & \shortstack{0.52\,$\pm$\,0.38\\[1pt] {\scriptsize (6/20)}} & \shortstack{0.55\,$\pm$\,0.46\\[1pt] {\scriptsize (9/20)}} & \shortstack{0.55\,$\pm$\,0.43\\[1pt] {\scriptsize (8/20)}} \\
\bottomrule
\end{tabular}
\end{table}

\begin{table}[t]
\centering\small
\renewcommand{\arraystretch}{1.15}
\setlength{\tabcolsep}{4pt}
\caption{\textbf{Mock Kitchen --- OOD-Pos.} SimVLA is best on task progress across all tasks, and best-or-comparable on binary success. Task progress $\pm$ s.d.\ over 20 trials; binary successes in parentheses.}
\label{tab:mock_ood_pos}
\begin{tabular}{lccccc}
\toprule
Model & Task 1 & Task 2 & Task 3 & Task 4 & Task 5 \\
\midrule
$\pi_{0.5}$ fine-tuned w/ 50 demos & \shortstack{0.40\,$\pm$\,0.38\\[1pt] {\scriptsize (4/20)}} & \shortstack{0.33\,$\pm$\,0.37\\[1pt] {\scriptsize (4/20)}} & \shortstack{0.40\,$\pm$\,0.37\\[1pt] {\scriptsize (4/20)}} & \shortstack{0.30\,$\pm$\,0.44\\[1pt] {\scriptsize (5/20)}} & \shortstack{0.25\,$\pm$\,0.41\\[1pt] {\scriptsize (4/20)}} \\
$\pi_{0.5}$ fine-tuned w/ 200 demos & \shortstack{0.45\,$\pm$\,0.39\\[1pt] {\scriptsize (5/20)}} & \shortstack{0.43\,$\pm$\,0.40\\[1pt] {\scriptsize (6/20)}} & \shortstack{0.48\,$\pm$\,0.35\\[1pt] {\scriptsize (4/20)}} & \shortstack{0.35\,$\pm$\,0.46\\[1pt] {\scriptsize (6/20)}} & \shortstack{0.28\,$\pm$\,0.44\\[1pt] {\scriptsize (5/20)}} \\
SimVLA (w/o pre-train) & \shortstack{0.23\,$\pm$\,0.34\\[1pt] {\scriptsize (2/20)}} & \shortstack{0.17\,$\pm$\,0.18\\[1pt] {\scriptsize (0/20)}} & \shortstack{0.13\,$\pm$\,0.17\\[1pt] {\scriptsize (0/20)}} & \shortstack{0.10\,$\pm$\,0.26\\[1pt] {\scriptsize (1/20)}} & \shortstack{0.25\,$\pm$\,0.38\\[1pt] {\scriptsize (3/20)}} \\
SimVLA (w/o SimDeploy) & \shortstack{0.45\,$\pm$\,0.36\\[1pt] {\scriptsize (4/20)}} & \shortstack{0.36\,$\pm$\,0.31\\[1pt] {\scriptsize (3/20)}} & \shortstack{0.43\,$\pm$\,0.38\\[1pt] {\scriptsize (5/20)}} & \shortstack{0.38\,$\pm$\,0.32\\[1pt] {\scriptsize (2/20)}} & \shortstack{0.50\,$\pm$\,0.32\\[1pt] {\scriptsize (4/20)}} \\
SimVLA & \shortstack{\textbf{0.58\,$\pm$\,0.41}\\[1pt] {\scriptsize (8/20)}} & \shortstack{\textbf{0.50\,$\pm$\,0.36}\\[1pt] {\scriptsize (6/20)}} & \shortstack{\textbf{0.52\,$\pm$\,0.37}\\[1pt] {\scriptsize (6/20)}} & \shortstack{\textbf{0.60\,$\pm$\,0.45}\\[1pt] {\scriptsize (10/20)}} & \shortstack{\textbf{0.68\,$\pm$\,0.41}\\[1pt] {\scriptsize (11/20)}} \\
\bottomrule
\end{tabular}
\end{table}

\begin{table}[t]
\centering\small
\renewcommand{\arraystretch}{1.15}
\setlength{\tabcolsep}{4pt}
\caption{\textbf{Mock Kitchen --- OOD-Vis.} SimVLA achieves the best performance on Tasks 1, 3, 4 and 5, whereas the $\pi_{0.5}$ model fine-tuned with 200 demonstrations performs best on Task 2. Relative to OOD-Pos, $\pi_{0.5}$ holds up better under OOD-Vis shifts, likely due to image augmentation during its fine-tuning. Task progress $\pm$ s.d.\ over 20 trials; binary successes in parentheses.}
\label{tab:mock_ood_vis}
\begin{tabular}{lccccc}
\toprule
Model & Task 1 & Task 2 & Task 3 & Task 4 & Task 5 \\
\midrule
$\pi_{0.5}$ fine-tuned w/ 50 demos & \shortstack{0.35\,$\pm$\,0.46\\[1pt] {\scriptsize (6/20)}} & \shortstack{0.37\,$\pm$\,0.39\\[1pt] {\scriptsize (5/20)}} & \shortstack{0.40\,$\pm$\,0.37\\[1pt] {\scriptsize (4/20)}} & \shortstack{0.50\,$\pm$\,0.49\\[1pt] {\scriptsize (9/20)}} & \shortstack{0.30\,$\pm$\,0.41\\[1pt] {\scriptsize (4/20)}} \\
$\pi_{0.5}$ fine-tuned w/ 200 demos & \shortstack{0.45\,$\pm$\,0.48\\[1pt] {\scriptsize (8/20)}} & \shortstack{\textbf{0.45\,$\pm$\,0.42}\\[1pt] {\scriptsize (7/20)}} & \shortstack{0.45\,$\pm$\,0.39\\[1pt] {\scriptsize (5/20)}} & \shortstack{0.53\,$\pm$\,0.47\\[1pt] {\scriptsize (9/20)}} & \shortstack{0.48\,$\pm$\,0.47\\[1pt] {\scriptsize (8/20)}} \\
SimVLA (w/o pre-train) & \shortstack{0.20\,$\pm$\,0.34\\[1pt] {\scriptsize (2/20)}} & \shortstack{0.13\,$\pm$\,0.15\\[1pt] {\scriptsize (0/20)}} & \shortstack{0.13\,$\pm$\,0.17\\[1pt] {\scriptsize (0/20)}} & \shortstack{0.18\,$\pm$\,0.29\\[1pt] {\scriptsize (1/20)}} & \shortstack{0.08\,$\pm$\,0.18\\[1pt] {\scriptsize (0/20)}} \\
SimVLA (w/o SimDeploy) & \shortstack{0.38\,$\pm$\,0.36\\[1pt] {\scriptsize (3/20)}} & \shortstack{0.29\,$\pm$\,0.28\\[1pt] {\scriptsize (2/20)}} & \shortstack{0.37\,$\pm$\,0.36\\[1pt] {\scriptsize (4/20)}} & \shortstack{0.43\,$\pm$\,0.34\\[1pt] {\scriptsize (3/20)}} & \shortstack{0.43\,$\pm$\,0.29\\[1pt] {\scriptsize (2/20)}} \\
SimVLA & \shortstack{\textbf{0.63\,$\pm$\,0.39}\\[1pt] {\scriptsize (9/20)}} & \shortstack{0.40\,$\pm$\,0.34\\[1pt] {\scriptsize (4/20)}} & \shortstack{\textbf{0.47\,$\pm$\,0.38}\\[1pt] {\scriptsize (6/20)}} & \shortstack{\textbf{0.58\,$\pm$\,0.41}\\[1pt] {\scriptsize (8/20)}} & \shortstack{\textbf{0.60\,$\pm$\,0.38}\\[1pt] {\scriptsize (8/20)}} \\
\bottomrule
\end{tabular}
\end{table}

\begin{table}[t]
\centering\small
\renewcommand{\arraystretch}{1.15}
\setlength{\tabcolsep}{4pt}
\caption{\textbf{Pretraining Ablation}, same tasks as submitted, 20 rollouts for each evaluation in mock kitchen-ID. Removing either SimVQA or TokSimAction lowers task progress on every task relative to full SimVLA; removing SimVQA degrades control mainly through less precise grasping. Task progress $\pm$ s.d.; binary successes in parentheses.}
\label{tab:ablation_mock_id}
\begin{tabular}{lccccc}
\toprule
Model & Task 1 & Task 2 & Task 3 & Task 4 & Task 5 \\
\midrule
SimVLA w/o SimVQA & \shortstack{0.43\,$\pm$\,0.47\\[1pt] {\scriptsize (7/20)}} & \shortstack{0.36\,$\pm$\,0.41\\[1pt] {\scriptsize (5/20)}} & \shortstack{0.37\,$\pm$\,0.40\\[1pt] {\scriptsize (5/20)}} & \shortstack{0.50\,$\pm$\,0.43\\[1pt] {\scriptsize (7/20)}} & \shortstack{0.45\,$\pm$\,0.43\\[1pt] {\scriptsize (6/20)}} \\
SimVLA w/o TokSimAction & \shortstack{0.40\,$\pm$\,0.38\\[1pt] {\scriptsize (4/20)}} & \shortstack{0.29\,$\pm$\,0.35\\[1pt] {\scriptsize (3/20)}} & \shortstack{0.42\,$\pm$\,0.37\\[1pt] {\scriptsize (5/20)}} & \shortstack{0.43\,$\pm$\,0.41\\[1pt] {\scriptsize (5/20)}} & \shortstack{0.50\,$\pm$\,0.43\\[1pt] {\scriptsize (7/20)}} \\
SimVLA w/o both & \shortstack{0.38\,$\pm$\,0.36\\[1pt] {\scriptsize (3/20)}} & \shortstack{0.26\,$\pm$\,0.35\\[1pt] {\scriptsize (3/20)}} & \shortstack{0.37\,$\pm$\,0.36\\[1pt] {\scriptsize (4/20)}} & \shortstack{0.48\,$\pm$\,0.44\\[1pt] {\scriptsize (7/20)}} & \shortstack{0.45\,$\pm$\,0.43\\[1pt] {\scriptsize (6/20)}} \\
SimVLA & \shortstack{\textbf{0.60\,$\pm$\,0.42}\\[1pt] {\scriptsize (9/20)}} & \shortstack{\textbf{0.50\,$\pm$\,0.36}\\[1pt] {\scriptsize (6/20)}} & \shortstack{\textbf{0.52\,$\pm$\,0.38}\\[1pt] {\scriptsize (6/20)}} & \shortstack{\textbf{0.55\,$\pm$\,0.46}\\[1pt] {\scriptsize (9/20)}} & \shortstack{\textbf{0.55\,$\pm$\,0.43}\\[1pt] {\scriptsize (8/20)}} \\
\bottomrule
\end{tabular}
\end{table}

\begin{table}[t]
\centering\small
\renewcommand{\arraystretch}{1.15}
\setlength{\tabcolsep}{1.5pt} % was 4pt; narrowed to fit the wider lab font
\caption{\textbf{Pantry and Real Home.} Task progress $\pm$ s.d.; binary successes in parentheses. Pantry is evaluated on Tasks 1, 4 and 5 with 20 trials each; Real Home on Tasks 1 and 3 with 10 trials each.}
\label{tab:pantry_realhome}
\begin{tabular}{lccccc}
\toprule
 & \multicolumn{3}{c}{Pantry \scriptsize(20 trials)} & \multicolumn{2}{c}{Real Home \scriptsize(10 trials)} \\
\cmidrule(lr){2-4}\cmidrule(lr){5-6}
Model & Task 1 & Task 4 & Task 5 & Task 1 & Task 3 \\
\midrule
$\pi_{0.5}$ fine-tuned w/ 50 Mock Kitchen demos & \shortstack{0.10\,$\pm$\,0.21\\[1pt] {\scriptsize (0/20)}} & \shortstack{0.08\,$\pm$\,0.18\\[1pt] {\scriptsize (0/20)}} & \shortstack{0.18\,$\pm$\,0.24\\[1pt] {\scriptsize (0/20)}} & \shortstack{0.20\,$\pm$\,0.26\\[1pt] {\scriptsize (0/10)}} & \shortstack{0.17\,$\pm$\,0.18\\[1pt] {\scriptsize (0/10)}} \\
$\pi_{0.5}$ fine-tuned w/ 200 Mock Kitchen demos & \shortstack{0.23\,$\pm$\,0.30\\[1pt] {\scriptsize (1/20)}} & \shortstack{0.23\,$\pm$\,0.30\\[1pt] {\scriptsize (1/20)}} & \shortstack{0.18\,$\pm$\,0.29\\[1pt] {\scriptsize (1/20)}} & \shortstack{0.35\,$\pm$\,0.24\\[1pt] {\scriptsize (0/10)}} & \shortstack{0.23\,$\pm$\,0.16\\[1pt] {\scriptsize (0/10)}} \\
SimVLA w/o pre-train & \shortstack{0.10\,$\pm$\,0.21\\[1pt] {\scriptsize (0/20)}} & \shortstack{0.10\,$\pm$\,0.26\\[1pt] {\scriptsize (1/20)}} & \shortstack{0.08\,$\pm$\,0.18\\[1pt] {\scriptsize (0/20)}} & \shortstack{0.10\,$\pm$\,0.21\\[1pt] {\scriptsize (0/10)}} & \shortstack{0.10\,$\pm$\,0.16\\[1pt] {\scriptsize (0/10)}} \\
SimVLA & \shortstack{\textbf{0.60\,$\pm$\,0.38}\\[1pt] {\scriptsize (8/20)}} & \shortstack{\textbf{0.53\,$\pm$\,0.47}\\[1pt] {\scriptsize (9/20)}} & \shortstack{\textbf{0.58\,$\pm$\,0.44}\\[1pt] {\scriptsize (9/20)}} & \shortstack{\textbf{0.70\,$\pm$\,0.42}\\[1pt] {\scriptsize (6/10)}} & \shortstack{\textbf{0.47\,$\pm$\,0.42}\\[1pt] {\scriptsize (3/10)}} \\
\bottomrule
\end{tabular}
\end{table}

\end{document}